\documentclass[preprint,12pt,authoryear]{elsarticle}

\usepackage{graphicx}
\usepackage{xcolor}
\usepackage{amssymb}
\usepackage{footmisc}
\usepackage{tablefootnote}
\usepackage{amsmath}
\usepackage{lineno}
\usepackage[flushleft]{threeparttable}
\usepackage{subcaption}

\journal{Journal}

\begin{document}

\begin{frontmatter}



\title{High-Precision Geosteering via Reinforcement Learning and Particle Filters}


\author[uis]{Ressi Bonti Muhammad\corref{cor}}
\ead{ressi.b.muhammad@uis.no}
\author[stanford]{Apoorv Srivastava\corref{cor}}
\ead{apoorv1@stanford.edu}
\author[norce]{Sergey Alyaev}
\author[uis]{Reidar Brumer Bratvold}
\author[stanford]{Daniel M. Tartakovsky}

\affiliation[uis]{organization={University of Stavanger},
            addressline={Kjell Arholms gate 41}, 
            city={Stavanger},
            postcode={4021}, 
            country={Norway}}
            
\affiliation[stanford]{organization={Stanford University},
            addressline={450 Jane Stanford Way}, 
            city={Stanford},
            postcode={CA 94305}, 
            country={USA}}
            
\affiliation[norce]{organization={NORCE Norwegian Research Centre},
            addressline={Nygårdsgaten 112}, 
            city={Bergen},
            postcode={5008}, 
            country={Norway}}
            
\newcommand{\todo}[1]{{\color{red} TODO: #1}}
\newcommand{\TODO}[1]{\todo{#1}}

\cortext[cor]{Corresponding authors:}

\begin{abstract}
Geosteering, a key component of drilling operations, traditionally involves manual interpretation of various data sources such as well-log data. This introduces subjective biases and inconsistent procedures. 
Academic attempts to solve geosteering decision optimization with greedy optimization and Approximate Dynamic Programming (ADP) showed promise but lacked adaptivity to realistic diverse scenarios. 
Reinforcement learning (RL) offers a solution to these challenges, facilitating optimal decision-making through reward-based iterative learning. 
State estimation methods, e.g., particle filter (PF), provide a complementary strategy for geosteering decision-making based on online information. We integrate an RL-based geosteering with PF to address realistic geosteering scenarios. Our framework deploys PF to process real-time well-log data to estimate the location of the well relative to the stratigraphic layers, which then informs the RL-based decision-making process. We compare our method’s performance with that of using solely either RL or PF. Our findings indicate a synergy between RL and PF in yielding optimized geosteering decisions.

\end{abstract}


\begin{highlights}
\item Integrated Approach for Geosteering: Presented a unified framework that combines Reinforcement Learning (RL) and the Particle Filter (PF) to optimize geosteering decisions.

\item Diverse Decision Criteria: Introduced three decision-making methods that leverage RL, PF, and their synergy, catering to different geosteering contexts, from direct well-log data utilization to state estimate-driven decisions.

\item Realistic Scenario Analysis: Applied the proposed methods to realistic geosteering decision-making scenarios, demonstrating their applicability and effectiveness in real-world contexts.
\end{highlights}


\end{frontmatter}


\newcommand{\todo}[1]{{\color{red} TODO: #1}}
\newcommand{\TODO}[1]{\todo{#1}}

\section{Introduction}
\label{sec:intro}
Geosteering is a sequential decision-making problem \citep{ALYAEV2021}, where a decision-maker faces a series of interdependent decisions over time. Optimized decision-making in geosteering is crucial for maximizing the wellbore length within the reservoir. Traditionally, geosteering decisions have been guided by human expertise and intuitions, often relying on manual interpretation of data sources such as well-log data measured using logging-while-drilling (LWD) tools and prior geological models. While this method has yielded valuable results, it is inherently subjective and does not follow a logically consistent procedure for making high-quality decisions \citep{Kullawan2014, ALYAEV2021, yasaman, amine2021}. 

Automated geosteering represents a significant leap forward, as evidenced by Denisenko et al. (2020), showing that artificial intelligence (AI) can provide geological interpretations that closely align with the assessments of expert geologists. Building upon our previous contribution to automate geosteering decision-making, we are now extending our study to include the integration of automated interpretation (or state estimation) algorithm. This integration is designed to improve the reliability and efficiency of our automated geosteering method, ensuring optimal well placement and maximized reservoir contact.

\subsection{Decision Optimization Methods in Geosteering}\label{decopt}
The pursuit of a more consistent and efficient decision optimization method for geosteering has prompted the exploration of alternative methods. Greedy optimization was used for decision-making by \citet{Chen2014} and \citet{Kullawan2014-2}. 
This method involves selecting locally optimal decisions or those that offer the greatest immediate gain at each stage of the decision-making process. However, relying solely on immediate gains without considering future decisions and learning in sequential decision-making scenarios will, in general, lead to sub-optimal decisions, which, in turn, will lead to sub-optimal value creation \citep{Kullawan2016, KULLAWAN201890}.

To overcome the limitations of greedy optimization, \citet{KULLAWAN201890} introduced Approximate Dynamic Programming (ADP). Unlike greedy optimization, ADP incorporates future decisions and learning in addition to immediate gains. 
This enables proactive decision-making, which is optimal within the chosen discretization \citep{Alyaev2018}. 
Thus, ADP ensures globally optimized decisions, leading to maximum value creation for sequential decision-making scenarios. A study by \citet{Kullawan2016} demonstrated that ADP outperforms greedy optimization regarding the objective function value when tested on various geosteering decision-making scenarios. However, ADP is not without limitations. The computational requirements associated with implementing ADP may render it unsuitable for real-time decision-making scenarios where time and cost are crucial factors. Furthermore, ADP is typically designed for a specific scenario, making it challenging to adapt to different scenarios \citep{muhammad2023optimal}. \citet{ALYAEV2019} introduced a simplified 'naive-optimistic' \citep{Alyaev2018} version of ADP, which increases the method adaptability but loses the proven optimality of ADP.

Reinforcement learning (RL) has emerged as a promising method for solving sequential decision-making scenarios while addressing the limitations associated with ADP. \citet{muhammad2023optimal} has demonstrated the application of RL in optimizing geosteering decisions. The "RL-based geosteering" method introduced by the study offers a promising approach for optimizing decisions in real-time geosteering scenarios.
RL refers to a method that trains an agent to make decisions based on rewards or penalties received from a decision-making environment. The agent develops optimal decision-making strategies through iterative learning from interactions with the environment. The study demonstrated that RL produces results comparable to ADP for the scenarios from \citet{Kullawan2014-2} and \citet{KULLAWAN201890}. 
At the same time, RL offers flexibility and computational efficiency. 

\citet{muhammad2023optimal} introduced two RL-based geosteering methods, both of which use distance to reservoir boundaries as their primary decision-making criterion. The main difference lies in the type of distance to reservoir boundaries used in each method. The first method uses look-ahead\footnotemark{}\footnotetext{Look-ahead refers to a location ahead of the current position of the drillbit.} distance to reservoir boundaries estimates from a Bayesian inference method. On the other hand, the second method relies on direct\footnotemark{}\footnotetext{Direct refers to value yielded without any state estimation method.} and current\footnotemark{}\footnotetext{Current is mentioned only once here. If look-ahead is not mentioned, the data is measured at the drillbit.} distance to reservoir boundaries measurements. The study concluded that both methods had identical performance in solving geosteering scenarios. However, the second method excels in computational efficiency as it learns the optimized decision-making strategy without relying on state estimation methods.

\subsection{State Estimation Methods in Geosteering}
\label{sec:state}
All the decision optimization methods described in Section~\ref{decopt} share a common feature: they either require or offer the option to use state estimation methods. This highlights the important role of state estimation methods in providing estimates of relevant geological characteristics as inputs to the decision optimization methods. When applied in a geosteering context, it is important for the state estimation method to provide accurate estimates while maintaining computational efficiency.

Deterministic state estimation methods, such as gradient descent, Gauss-Newton, and Levenberg-Marquardt, are widely used in geosteering \citep{bakr2017fast,sviridov2014new,thiel20182d,wu2018new}. The methods are generally gradient-based and involve minimizing the error term between real-time measurements and the responses from the forward model. Due to their computational efficiency, deterministic state estimation methods are well-suited for real-time geosteering applications. However, a notable drawback is their susceptibility to the local minimum problem and sensitivity to the initial guess \citep{jin2020}. Furthermore, these methods lack the ability to provide uncertainty quantification \citep{jahani2022}.

To address these challenges, considerable efforts have been dedicated to developing improved state estimation methods for real-time geosteering scenarios. Several studies explored probabilistic state estimation methods. 
Ensemble-Kalman-filter- \citep{ALYAEV2019} and ensemble-smoother-type \citep{rammay2022probabilistic,jahani2022,jahani2023enhancing} methods are extremely fast probabilistic methods that update a subsurface state represented as an ensemble of realizations. However, these methods can be sensitive to prior selection and require special to alleviate multi-modality in the solution \cite{rammay2022probabilistic}, which might be a constraining factor in a general operation.

More general, but slower, methods include the Markov chain Monte Carlo \citep{LU2019}, hybrid Monte Carlo \citep{SHEN2018}, and sequential Monte Carlo or the Particle Filter (PF) method \citep{Veettil20}. 
Unlike the deterministic state estimation methods, these probabilistic methods offer the advantage of exploring multiple state estimates that closely align with the true geological characteristics. However, their overall computational requirements are higher than the deterministic state estimation methods.

Other studies in state estimation methods for geosteering revolve around using deep neural networks (DNNs).
Most of the studies build fast determenistic estimators. \citep{puzyrev2021inversion} demonstrated the capability of deep convolutional neural networks to accurately perform 1D inversion of electromagnetic survey data, specifically enhancing real-time exploration by estimating the subsurface resistivity distribution.
\citet{jin2020} introduced a physics-driven deep learning approach that combines a forward physical model with a convolutional neural network. \citep{shahriari2020deep} demonstrated the effectiveness of deep learning algorithms in the rapid inversion of borehole resistivity measurements for real-time geosteering, while their subsequent work \citep{shahriari2021error} emphasized the importance of error control and the strategic selection of loss functions within DNNs to ensure accuracy and reliability.
\citet{alyaev2022} proposed a probabilistic mixture density DNN trained with "multiple-trajectory-prediction" loss functions, resulting in a DNN capable of delivering several possible state estimates and their probabilities within milliseconds. The DNN was later tested for sequential predictions \citep{alyaev2022sequential}.

In this study, we use PF as our state estimation method. \citet{Veettil20} demonstrated the effectiveness of PF in stratigraphic-based geosteering scenarios. This particular scenario refers to estimating well location relative to the stratigraphic layers of the reservoir. The stratigraphic-based geosteering scenarios align well with the RL-based geosteering method proposed by \citet{muhammad2023optimal}, as the state estimates derived from PF can be used to estimate the distance to reservoir boundaries. Moreover, PF is a widely adopted state estimation method known for its relatively straightforward implementation. 

\subsection{Contribution}
In this study, our contribution lies in combining the core idea behind the RL-based geosteering methods and proposing an improved method capable of addressing realistic geosteering scenarios. We aim to align the RL-based geosteering method with the widely used decision-making process in geosteering practices. Specifically, we achieve this by integrating a state estimation method (PF) that uses real-time well-log data to estimate well location relative to the stratigraphic layer of the reservoir. The relative well location can be used to estimate the distance to reservoir boundaries. By incorporating the estimates (the first RL-based geosteering method) of distance to reservoir boundaries (the second RL-based geosteering method) as the primary decision-making criterion, we can create a robust and suitable framework for making decisions in realistic geosteering scenarios.

Additionally, we propose another method for real-time decision-making scenarios using RL, where we directly use the well-log data as the primary criterion. This method leverages RL independently of PF. As an alternative to the RL-based methods, we present a rule-based decision-making method that uses the outputs from PF to inform its decisions. 
Therefore, our main contribution includes three decision-making methods: one that solely relies on RL, another that uses PF independently of RL, and a third that combines RL and PF.

To measure the performance of these three methods, we apply them to a geosteering decision-making context that uses functions from \citet{randomstrat} to generate the reservoir boundaries randomly. The well-log data in this scenario is taken from the Geosteering World Cup 2020 \citep{log, alyaev2022}.
Compared to the geosteering decision-making scenarios optimized in \citet{muhammad2023optimal}, this particular scenario presents a more realistic scenario, which aligns with our study objective.

\section{Methodology}
In this section, we describe the methodologies supporting our study, centered around two key elements: PF and RL. The PF serves as the state estimation method, adeptly estimating the states of the reservoir boundaries through probabilistic representations. Complementing this, RL is used to optimize decision-making by evaluating actions in terms of their expected rewards, thereby facilitating more informed decisions in dynamic and uncertain environments.
\subsection{Particle Filter}
\label{PF}
The state estimation problem, encountered frequently in engineering applications, is addressed either through optimization-based variational methods summarized in~\citet{varDAReview} or via methods based on Bayesian inference, such as Kalman filter (and its many versions) and PF, outlined by~\citet{BayesianDAReview}. Bayesian techniques, in contrast to variational methods, offer ways to quantify the uncertainties in the estimated state. Such estimates are essential to decision-making processes under uncertainties; consequently, Bayesian methods are used for state estimation problems in this work. 

The probabilistic nature of the Bayesian framework quantifies the uncertainties by expressing the states and observations in terms of their respective probability distribution functions (PDFs). For a dynamical system under Markovian assumptions, at time $t$, the conditional distribution $f_{S_t|o_t}(s_t)$ on the state $s_t$ given observations $o_t$ is updated sequentially using the Bayes rule as \citep{PFristic},
\begin{align}
    \label{eqn:sequentialPDFUpdate}
    f_{S_t|o_t} \propto f_{o_t|S_t}f_{S_{t}|S_{t - 1}}f_{S_{t - 1}|o_{t - 1}},
\end{align}
where, $f_{o_t|S_t}(s_t)$, termed likelihood density, signifies the probability of observing $o_t$ given that the state is $s_t$, the PDF $f_{S_{t}|S_{t - 1}}$ encapsulates the stochasticity in model dynamics, and $f_{S_{t - 1}|O_{t - 1}}$ is the equivalent of $f_{S_t|o_t}$ from the previous step.


Kalman filter-based methods are typically characterized by Gaussian assumptions on the PDFs involved in Equation~\ref{eqn:sequentialPDFUpdate}, whereas PF is known to accommodate arbitrary distributions. To avoid any assumptions on the distributions involved, PF is employed in the results presented.

The assumption of PF and the algorithm used in this study, as elaborated in~\citet{PFristic} and ~\citet{SRIVASTAVA2023112499}, along with its implementation in the context of the decision-making in geosteering problems, are discussed next.

\textbf{Particle Filter.}
The PF approximates the PDFs involved in Equation~\ref{eqn:sequentialPDFUpdate} through weighted discrete samples. The PDF $f_{S_t|O_t}(s)$ is approximated by
\begin{align}
    \label{eqn:DiscreteApproxPF}
    f_{S_t|O_t}(s) \approx \sum_{i = 1}^{N_\text{par}} w^j_t \delta(s - s^i_t), \quad w^i_t \propto \frac{f_{S_t|O_t}(s^i)}{q_{S_t|O_t}(s^i)},
\end{align}
where, $\delta(s)$ is the Dirac delta function, $\{s^1_t,\dots,s^{N_\text{par}}_t\}$ are discretely sampled points, called particles, and $\{w^1_t,\dots,w^{N_\text{par}}_t\}$ are the associated weights such that $\sum_{i = 1}^{N_\text{par}} w^i_t = 1$. The pairs $\{s^i_t, w^i_t\}_{i = 1}^{N_\text{par}}$ are obtained through importance sampling, in which the particles $s^i_t$ are sampled from $q_{S_t|O_t}(s)$, any known density with the same support as $f_{S_t|O_t}(s)$, called importance density. 

Suitably choosing the importance density $q_{S_t|O_t}$  results in a sequential weight update rule,
\begin{align}
    \label{eqn:weightUpdatePF}
    w^i_t  \propto f_{O_t|S_t}(s^i) w^i_{t - 1}.
\end{align}
The likelihood density $f_{O_t|S_t}$ in the update rule (Equation~\ref{eqn:weightUpdatePF}) is designed to capture the observation model and error and is specific to the problem under consideration.

To eschew the degeneracy phenomenon in which most of the particles assume negligible weight, the particles are resampled after a fixed number of estimation steps. Resampling serves to eliminate the less weighted particles and duplicate the highly weighted particles, and in doing so, increases the density of particles in the high probability regions of the distribution, improving the PDF approximation.

\textbf{Algorithm.} 
The PF is initiated with a known initial PDF $f_{S_0} \equiv f_{S_0|O_0}$, where $O_0 = \phi$ is a null set and indicates the absence of any observations. The particles $\{s^1_0,\dots,s^{N_\text{par}}_0\}$ are sampled using the importance density $q_{S_0} = f_{S_0}$, and, using the definition of weights from Equation~\ref{eqn:DiscreteApproxPF}, are assigned equal weights, $w^i_0 = 1/N_\text{par}$. 

At any time $t \geq 1$, given the PDF $f_{S_{t - 1}|O_{t - 1}}$, approximated using $\{s^i_{t - 1}, w^i_{t - 1}\}_{i = 1}^{N_\text{par}}$ via Equation~\ref{eqn:DiscreteApproxPF}, the particles $\{s^i_{t - 1},\dots,s^{N_\text{par}}_{t - 1}\}$ are propagated through the forward model, resulting in states $\{s^1_t,\dots,s^{N_\text{par}}_t\}$. Bequeathed with observation $o_t$ at time $t$, the approximation of the updated PDF $f_{S_t|O_t}$ is specified by $\{s^i_t, w^i_t\}_{i = 1}^{N_\text{par}}$, where the weights $w^i_t \ (i = 1, \dots, N_\text{par})$ are obtained using Equation~\ref{eqn:weightUpdatePF}. 

To alleviate the aforementioned degeneracy phenomenon in PF, particles are resampled periodically. Resampling involves drawing $N_\text{par}$ independent samples from the discrete PDF defined by $\{s^i_{t - 1}, w^i_{t - 1}\}_{i = 1}^{N_\text{par}}$. All resampled particles, being i.i.d. samples, are equally weighted, i.e., $w^i_t = 1/N_\text{par}, i = 1, \dots, N_\text{par}$ after resampling.

\textbf{Implementation.} 
In the present work, PF is used to extract a probabilistic description of the well location with respect to the reservoir boundaries using well-log data. At any given location $x$ along the reservoir, the position of the reservoir boundaries $b_x$ is defined as the state, i.e., $s = b_x$, and the gamma-ray log observations from the drilled well $g_x$ are used as the observations $o$. Consequently, the observation operator for the problem can be summarized as
\begin{align}
    g_x = f(b_x),
\end{align}
with the objective of estimating the reservoir boundaries $b_x$ using the gamma-ray log observations from the drilled well $g_x$ and an offset well $f(x)$. The offset well is drilled in the vicinity of the currently drilled well and is assumed to have similar characteristics.

The PF is initiated with uninformative uniform PDFs $f_{S_0|O_0} \sim U(a, b)$ over a range of possible states to reflect the absence of prior knowledge, and the well-log data is assimilated after every advancement of fixed length $\Delta x = 10$ ft. The likelihood function $f_{O_t|S_t}$ critical to PF is designed to be a Gaussian distribution
\begin{align}
    \label{eqn:likelihoodFunction}
    f_{O_t|S_t}(s) \equiv \mathcal{N}(s; \mu, \sigma)
\end{align}
with mean $\mu = O_t$ and standard deviation $\sigma = 0.2$. The resampling procedure to avoid degeneracy is carried out before each decision point and midway between any two decision points. 

The $N_\text{par}$ particles obtained after assimilating the well-log data represent the possible values of the stratigraphic layer function with corresponding weights indicating their likelihood; such probabilistic representation is crucial for capturing the stochastic nature of the subsurface. We would like to note that though in this study, a single observation of well-log data at each step is used to estimate the stratigraphic layer function, it can easily be extended to multiple well-log observations by defining the likelihood function $f_{O_t|S_t}$ as a product of Gaussian densities from Equation~\ref{eqn:likelihoodFunction} corresponding to each observation.






\subsection{RL-based Geosteering}
\label{RL}
This subsection provides a more detailed description of the RL-based geosteering decision optimization method proposed by \citet{muhammad2023optimal}. This method is a fundamental basis upon which we will develop a more robust and practical decision optimization method designed to address realistic geosteering scenarios. We will describe the RL-based geosteering method, starting with its core concept of RL. Subsequently, we will discuss the specific RL algorithm used. Finally, we will conclude the section by exploring the implementation of this method in a geosteering scenario previously discussed in the study.

\textbf{Reinforcement learning.} RL is a decision optimization method for understanding and automating sequential decision-making scenarios \citep{sutton2018}. The term "automate" in RL arises from its ability to acquire optimal decision-making strategies through unsupervised and direct interaction with the decision-making scenario (environment). The interaction is made within a Markov Decision Process (MDP) system, a class of stochastic sequential decision processes used to model decision-making scenarios in discrete, stochastic, and sequential domains \citep{PUTERMAN1990331}.

\begin{figure}[ht]
    \centering
    \includegraphics[scale=0.4]{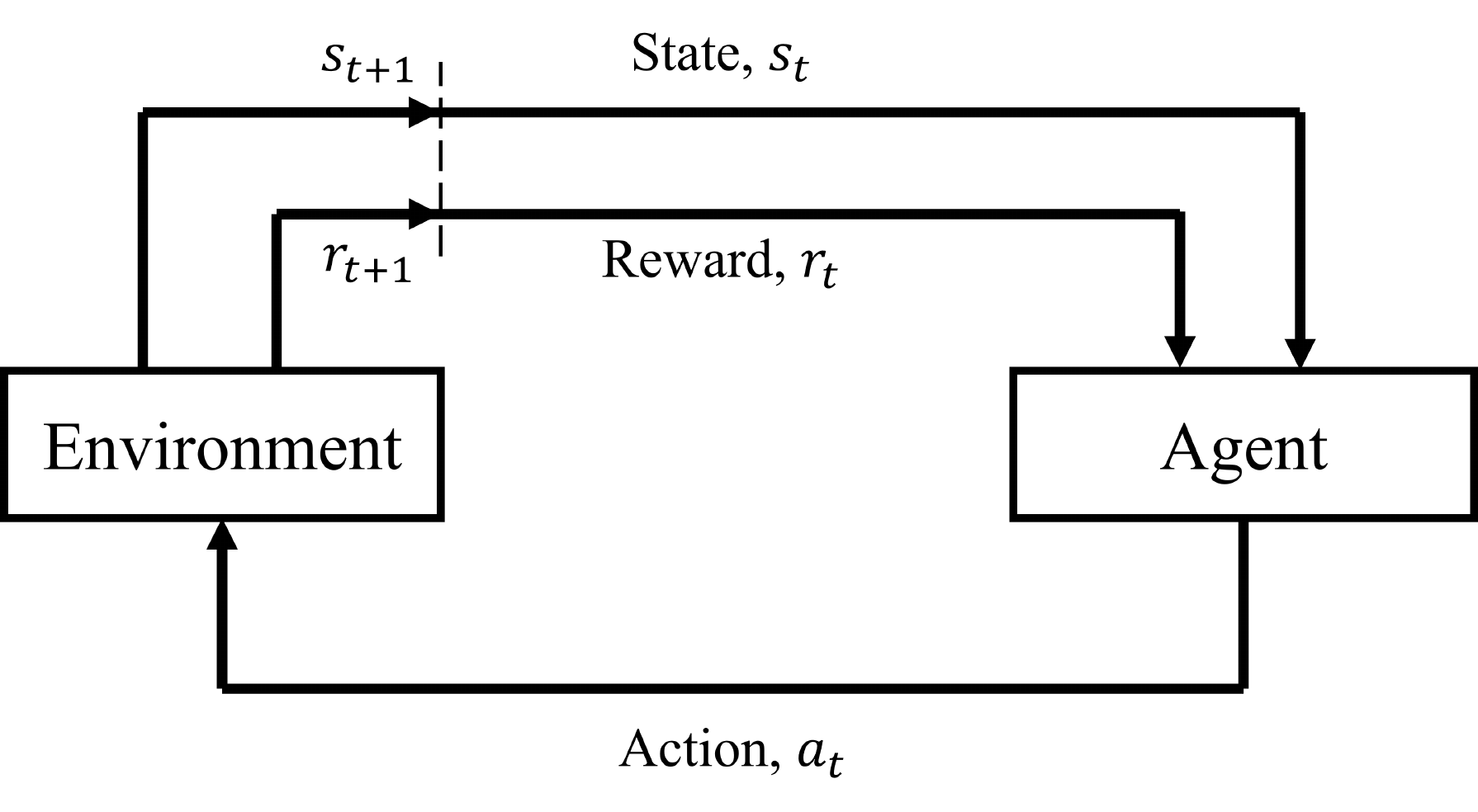}
    \caption{RL agent interaction with a decision-making environment \citep{muhammad2023optimal}. A decision-making agent interacts with a decision-making environment by receiving state $s_t$ and reward $r_t$ inputs, taking action $a_t$, and receiving feedback $s_{t+1}$ and $r_{t+1}$ in response}. 
    \label{fig:RL figure}
\end{figure}

The interaction is illustrated in Figure \ref{fig:RL figure}. During each iteration of the decision-making process, an agent takes action, $a_t$, based on the current state of the environment, $s_t$. The environment provides feedback to the agent in the form of a reward, $r_{t+1}$, while transitioning to a new state, $s_{t+1}$. This process is repeated, which allows the agent to adjust its decision-making strategy to maximize value creation.

The RL-based geosteering method uses an action-value-based algorithm to improve the decision-making capabilities of the agent. The primary objective of the method is to improve the understanding of the action-value function, which represents the expected value associated with taking a specific action in a given state. The agent can optimize its decision-making strategy by improving its understanding of this function, consistently selecting actions with the highest value.

\textbf{Algorithm.} Q-learning is one of the action-value-based algorithms designed to estimate the true action-value (Q-value) function. This method relies on the following update rule for Q-value:
\begin{equation}\label{eq:q-learning}
    Q(s_t,a_t) \leftarrow Q(s_t,a_t) + \alpha[r_{t+1} + \gamma max_{a'}Q(s_{t+1},a') - Q(s_t,a_t)].
\end{equation}
In this context, $Q(s_t,a_t)$ represents the current value assigned to perform a specific action $a_t$ in a given state $s_t$. The Q-value is updated by considering the immediate reward $r_{t+1}$, the maximum Q-value among all possible actions in the subsequent state $max_{a'}Q(s_{t+1},a')$, and the current Q-value. The learning rate $\alpha$ controls the magnitude of the update, while the discount factor $\gamma$ determines the relative importance of the next state values.

In Q-learning, Q-values are updated and stored in a tabular setting. The convergence criterion for Q-learning is to update all state-action pairs continuously \citep{sutton2018}. While this criterion is adequate for small-scale scenarios, it becomes challenging to apply in large-scale scenarios, particularly those with continuous state spaces. To address this limitation, \citet{Mnih2015-mf} proposed an alternative method called the Deep Q-Network (DQN) that approximates and stores Q-values using a deep neural network (DNN), replacing the tabular setting of the traditional Q-learning. However, including a DNN within an RL framework can lead to instabilities or divergent learning \citep{TDdivergence}. To effectively address these challenges, the DQN method illustrates the importance of employing mitigation methods, such as experience replay and target networks\footnotemark{}\footnotetext{More detailed definition of experience replay and target networks can be found in \citet{Mnih2015-mf}}, which are crucial for ensuring stability and convergent learning.

DQN learns by updating the parameter of the DNN through the minimization of the following loss function:
\begin{equation}\label{eq:dqn}
    L(\theta) = [y_t - Q(s_t,a_t;\theta_{t})]^2.
\end{equation}
In the given equation, $y_t$ represents the target Q-value, and $Q(s_t,a_t;\theta_{t})$ represents the predicted Q-value. The DNN is trained using stochastic gradient descent to update the network weights $\theta_{t}$ at each time-step $t$, resulting in improved estimates of the Q-values.

The RL-based geosteering method uses a DNN consisting of multiple hidden layers, where the input to the network is a state representation of the environment, and the outputs are Q-values for each possible action. The DNN architecture allows flexibility in adapting to different scenarios by adjusting the number of nodes in both the input and output layers. The configuration of the hidden layers is not strictly specified, but it impacts the outputted Q-values and the computational requirements. To provide an illustrative example, we will demonstrate the procedure of initializing DQN and its corresponding DNN for a geosteering decision-making scenario.

\textbf{Implementation.} We consider a geosteering scenario involving a horizontal well drilled in a thin, faulted, three-layered reservoir model. Our prior geological model provides information regarding the reservoir boundaries and fault information. We assume that the reservoir maintains a constant thickness and uniform properties throughout. 

A decision-making agent is involved in a decision-making process to optimize the length of the well within the reservoir. At each step of this process, the agent receives direct distance to reservoir boundaries measurements. The distance measurements in this scenario are assumed to be accurate. Afterward, the agent can make estimates about the look-ahead reservoir boundaries.

\citet{muhammad2023optimal} proposed two RL-based geosteering methods that demonstrated identical near-optimal efficacy in solving the geosteering decision-making scenario. In the decision-making process, the first method relies on look-ahead Bayesian estimates, while the second method uses direct distance to reservoir boundaries measurements. Despite their differences, both methods share similarities in leveraging direct distance to reservoir boundaries measurements. The second method directly uses these measurements for decision-making, while the first method uses them to update its prior geological model and generate probabilistic look-ahead estimates. 

The first method demonstrates how RL can optimize decisions when estimates are the primary decision-making criterion. In contrast, the second method highlights the capability of RL to make optimized decisions based solely on distance measurements. While either method can be selected, in this case, we choose the distance measurements as the state representation of the environment, along with some additional related information\footnotemark{}. The state representation serves as the input to the DNN.

After receiving the state representation, the agent chooses the optimal alternative (action) from five available alternatives. Following the steering decision, there is a subsequent conditional sidetrack decision, which can only be chosen if the well exits the reservoir. If selected, the well is directed back into the reservoir. In a conventional decision-making method, the steering and sidetrack decisions are typically treated as separate decision nodes, necessitating the use of two distinct DNNs within the DQN framework. However, in this specific context, we use a single DNN with six output nodes\footnotemark[\value{footnote}]\footnotetext{More in-depth explanation of the implementation of RL in this geosteering scenario can be found in \citet{muhammad2023optimal}.}.

Following each decision, the agent receives feedback in the form of a reward function. If not already defined, the reward function should align with the overall objective of the geosteering scenario. In this case, the objective is to maximize the length of the well within the reservoir (reservoir contact) while simultaneously minimizing operational costs. One possible reward function for this scenario is to adopt a common multi-objective decision analytic approach \citep{bratvoldbook}, where the objectives are weighted. Therefore, the reward function for this particular scenario can be written as follows:
\begin{equation}\label{eq:reward_ex2}
r = w_1\cdot v - w_2\cdot c.
\end{equation}
Here, $v$ represents the value given the location of the well, with a value of $v$ indicating that the well is located within the reservoir and 0 otherwise. The operating cost $c$ is the sum of the drilling cost and, if applicable, the sidetrack cost. The weight of objectives one and two are represented as $w_1$ and $w_2$, respectively.

The decision-making process is subsequently repeated as the environment transitions to a new state. Subsequently, the agent receives the updated state representation and chooses the next alternative. This entire process is repeated until the geosteering operation has been completed, which can be referred to as one episode. The agent may need to be trained over several thousand episodes to improve the decision-making strategy. This training procedure allows the agent to learn from the feedback received through the reward function, continuously improving its understanding of the optimal Q-values. 

In summary, implementing the RL-based geosteering method involves several steps. Firstly, the DNN is initialized by defining the state representation of the environment as the input and specifying the number of alternatives as the output. This ensures compatibility between the network and the geosteering scenario. Secondly, defining a reward function that aligns with the overall objective is essential if it is not initially available. This step ensures that the RL agent learns to maximize the desired objectives effectively. Finally, the training procedure can be initiated, allowing the RL agent to learn and improve its geosteering decision-making strategy over time by interacting with the environment.

\section{Proposed Method}\label{proposed}
This section will describe three methods used to optimize geosteering decisions. Each method represents a different approach: one uses RL independently of PF, another relies on PF independently of RL, and the third combines both methods. These methods ensure a comprehensive evaluation of what RL and PF bring to optimize realistic geosteering decision-making scenarios.

For methods that use RL, we will propose the state representation that suits each method most effectively. The remaining aspects of the methods will remain unchanged, as described in Section \ref{RL}. On the other hand, using PF or any other state estimation method without a decision optimization method is generally not very helpful in a geosteering decision-making context. Therefore, we will introduce a rule-based decision-making method that utilizes PF outputs as a critical input to inform and guide the decision-making process.

\subsection{RL State Representations}\label{rlstaterep}
\begin{figure}[ht]
    \centering
    \includegraphics[scale=0.735]{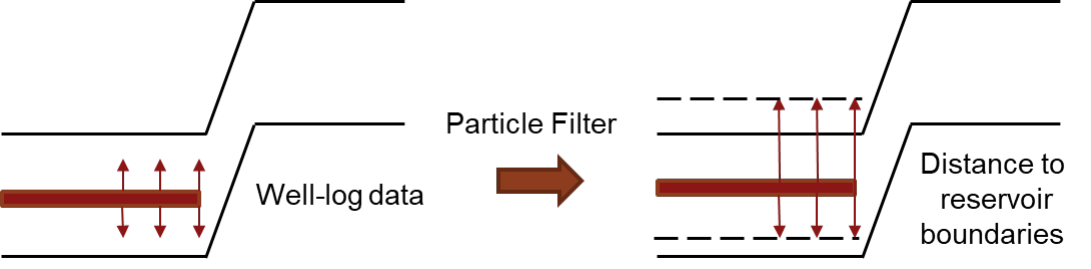}
    \caption{Illustration of using well-log data as inputs for PF to estimate the distance to reservoir boundaries. The solid black lines show the true reservoir boundaries, while the dashed black lines show the estimated boundaries derived from PF. The red lines show the trajectory of the drilled well, while the red arrows show the decision-making criterion. The figure on the left illustrates the RL-Log method, while the complete illustration refers to the RL-Estimation method.} 
    \label{fig:rllog}
\end{figure}

Before describing the representations, it is important to establish that the RL-based geosteering method is extended to align with the common practices in geosteering. Prior to each decision, we receive well-log data, which can be used directly, or as input to PF. Figure \ref{fig:rllog} illustrates the process of receiving well-log data before using it as inputs to PF to estimate the distance to reservoir boundaries. The details of the figure will be described alongside each corresponding method. We propose the following state representations that will serve as inputs to our DNN:

\textbf{Direct well-log data measurements.} The state representation is related to the second RL-based geosteering method. It emphasizes the utilization of direct measurements rather than relying on state estimation methods for representing the state. However, the main difference lies in utilizing well-log data commonly used during geosteering operations, such as gamma-ray, porosity, and resistivity logs. By using these widely available measurements, we ensure the practicality of the representation. We refer to this method as "RL-Log."

RL-Log state representation is shown in the left-hand side of Figure \ref{fig:rllog}. The solid black lines show the true\footnotemark{}\footnotetext{True refers to the actual reservoir, used as a benchmark to compare and validate the estimated reservoir boundaries} reservoir boundaries, where the red line shows the drilled well, and the red arrows show the surrounding locations from which the well-log data is obtained. The RL-Log method uses this data exclusively to optimize decision-making processes.

One advantage of RL-Log, or utilizing direct measurements as the state representation in general, is its computational efficiency (Muhammad et al., 2023). By avoiding the need for state estimation methods, the computational requirements are primarily determined by the complexity of the geosteering scenarios and the configuration of the DNN. However, RL-Log comes with a potential drawback. The subsurface is highly variable, leading to variations in well-log data across different geosteering scenarios. Consequently, the decision-making strategies derived from RL-Log may be scenario-specific or limit their applicability to scenarios used during the training procedure. 


\textbf{Outputs from PF.} This state representation is related to the first and second RL-based geosteering methods, as it incorporates a method to estimate the distance to reservoir boundaries and uses the distance to reservoir boundaries as its primary decision-making criterion. This method aligns with the common practice in geosteering decision-making, which has been emphasized throughout this study. We refer to this method as "RL-Estimation."

The entirety of Figure \ref{fig:rllog} describes RL-Estimation state representation. After receiving well-log data, the PF is used to process the data, resulting in an estimated distance to the reservoir boundaries. These estimated reservoir boundaries are indicated by the dashed black lines, and the corresponding distances from the drilled well, from which the RL-Estimation optimized the decisions, are indicated by the red arrows.

One advantage of RL-Estimation, specifically using distance to reservoir boundaries measurements (or estimates in this context) as the state representation, lies in its robustness and flexibility (Muhammad et al., 2023). The core premise of this method is that within the geosteering context, distance to reservoir boundaries serves as a more adaptable decision-making criterion than direct well-log data. Hence, even when applied to scenarios with different well-log data, the strategies remain applicable as the distance to reservoir boundaries can be estimated using the well-log data. 

However, RL-Estimation has a potential drawback, which is the increased computational requirements due to using a state estimation method. The requirement to run this method before making decisions significantly raises the computational requirements compared to RL-Log. This increment in computational requirements is particularly noticeable during the training procedure, where the requirements are influenced by the number of training episodes. Therefore, it is crucial to consider this factor when assessing the feasibility and practicality of implementing RL-Estimation, especially in realistic geosteering scenarios.

\subsection{Rule-based Decision-Making Method}
The method involves making na\"{i}ve estimates regarding the potential look-ahead depth of reservoir boundaries. Unlike the RL-based geosteering method, which can use any available information through its DNN for decision-making, traditional decision optimization methods usually depend on the look-ahead estimates to make optimized decisions. In this context, the rule-based decision-making method can be described as follows:

After completing a PF step, we select $X$ particles that best represent the likely reservoir boundaries based on the particles PDF after evaluating the observed well log data. Each best particle is expected to provide a good estimate of the true reservoir boundaries (due to their alignment with the observed well log data). This makes them a solid initial reference point for look-ahead estimates. We propose a na\"{i}ve look-ahead estimation method that uses the same state transition function used during the PF step to generate the look-ahead estimates from these particles. The state transition function is executed $n$ times, where $n$ represents the number of discrete points between two decision points in the geosteering scenario. However, similar to its utilization in the PF step, the state transition function generates these estimates in a stochastic manner, making the estimated look-ahead reservoir boundaries inherently have higher uncertainty than the particles.

\begin{figure}[ht]
    \centering
    \includegraphics[scale=0.735]{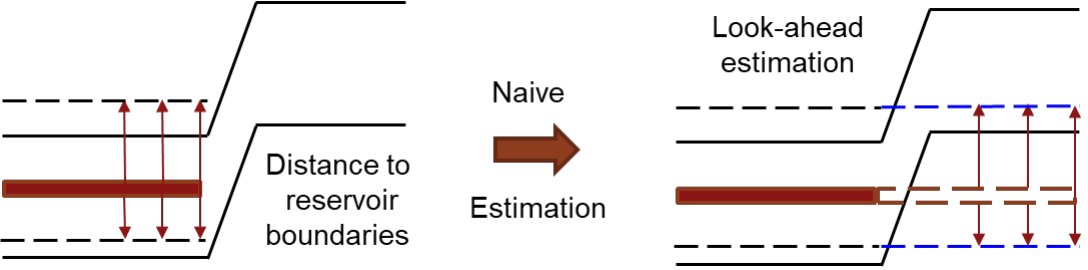}
    \caption{Illustration of the rule-based decision-making method. It shows the use of a PF particle as an initial reference to generate look-ahead estimates. The estimates are shown as dashed blue lines. Red arrows show distances to estimated look-ahead boundaries from the projected well, which are shown as dashed blue lines.} 
    \label{fig:naive}
\end{figure}

Figure \ref{fig:naive} illustrates the use of a particle from PF, similar to the one used by RL-Estimation in Figure \ref{fig:rllog}, as the initial reference point. The particle is the basis for generating look-ahead estimates, indicated by the dashed blue lines. These dashed blue lines illustrate the stochastic nature of look-ahead estimates, which may lead to significant deviations from the true reservoir boundaries. The red arrows indicate the distances from the projected well, represented by the dashed red lines, to the estimated look-ahead reservoir boundaries. The calculation of the projected well is based on a single alternative, and this process is repeated for all available alternatives, as will be described further below.

Once the estimation step is completed, the subsequent step is a decision-making step that involves iterating through all alternatives and selected particles to calculate the expected reward for each alternative. The expected reward for each alternative, $R$, is computed as the weighted sum of the rewards for all particles and discrete points:
\begin{equation}
    R = \sum_{i=1}^{X} w_i \cdot \sum_{j=1}^{N} r_{ij}. 
\end{equation}
Here, $w_i$ represents the normalized weight assigned to the $i$-th particle, indicating its significance in the estimation process. The normalized weight for each particle is calculated using the total weight of the set of $X$ particles. The reward function, $r_{ij}$, is obtained based on the $i$-th particle at the $j$-th discrete point, with $N$ equals to the number of discrete points of the scenario. 

The rule-based decision-making method adapts the following reward function:
\begin{equation}
    \label{eq:reward_naive}
    r = -4(\frac{d_{min}}{h} - 0.5)^2 + 1 
\end{equation}
where $r$ represents the reward function for one discrete point. The term $d_{min}$ corresponds to the minimum value between the distances from the drill bit to the top and bottom reservoir boundaries. Furthermore, $h$ shows the thickness of the reservoir. The reward function reaches its maximum value (1) when the well is optimally placed in the middle of the reservoir ($d_{min}/h = 0.5$). However, it is important to note that the function is unbounded for its minimum value, as it is specifically designed to penalize situations where the well is located far outside the reservoir boundaries. 

Following the computation of expected rewards for each alternative, the decision-making process concludes by selecting the best alternative, corresponding to the one with the highest expected reward. Opting for the alternative that offers the highest immediate expected reward mirrors the greedy optimization method. Consequently, the rule-based decision-making method is likely to yield locally optimized decisions.

\section{Numerical Example}
In this section, we will describe the geosteering decision-making scenario used for comparing the performance of the three proposed methods. Furthermore, we will outline the implementation and training procedures for the RL-Log and RL-Estimation methods. Finally, we will describe the evaluation procedure applied to all three proposed methods in this study.

\subsection{Geosteering Decision-Making Scenario}\label{scenario}
In this geosteering decision-making scenario, we focus on a horizontal well drilled in a faulted three-layered reservoir model. The reservoir is assumed to maintain a constant thickness of $h = $ 20 ft and uniform quality. Primary uncertainties in this scenario lie in the depths of the reservoir boundaries, which are randomly generated using functions from \citet{randomstrat}. The horizontal section of the reservoir boundaries is discretized into $N =$ 320 discrete points, each separated by 10 ft. The number of faults in this scenario is limited to three, with each fault separated by at least 100 points (1000 ft).

\begin{figure}[ht]
    \centering
    \includegraphics[scale=0.5]{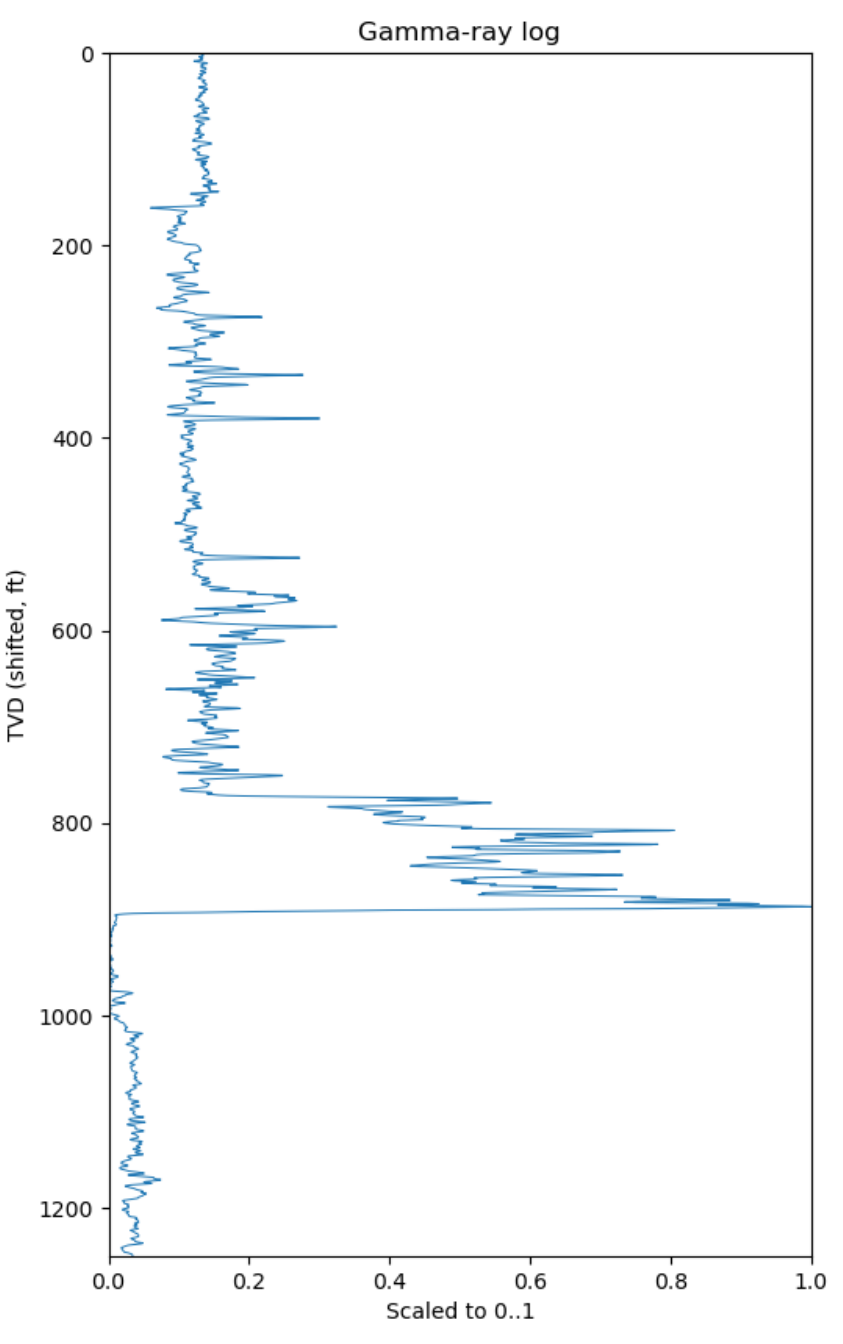}
    \caption{Gamma-ray log data based on \citet{log,alyaev2022}.} 
    \label{fig:log}
\end{figure}

As the well is drilled past each decision point, real-time well-log data\footnotemark{}\footnotetext{Data measurement is assumed to provide perfect information, but the model can incorporate uncertainties in the measurements if needed.} is measured using LWD tools. The well-log data used for this study is the gamma-ray log data \citep{log} used during the semifinal of the Geosteering World Cup 2020 \citep{amine2021}. The full gamma-ray log data used in this study is illustrated in Figure \ref{fig:log}. The data is shifted to a desired true vertical depth (TVD) and scaled to be between 0 and 1. The gamma-ray log data can be used as inputs to PF, with its outputs becoming the primary criterion for decision-making. Alternatively, the gamma-ray log data can be directly used as the primary decision-making criterion.

In this case, the decision-making agent makes well-steering decisions by adjusting the well inclination every $n = $ 32 discrete points, resulting in 10 decision points. At each decision point, the agent has 11 alternatives for adjusting the inclination, constrained between $-5$ and $5$ degrees with an increment of $1$ degree. To further restrict the steering decisions, we include another constraint where the inclination is bounded between 86 and 94 degrees, with values above 90 indicating the well is moving towards deeper depths. Additionally, the dogleg severity (DLS) constraint of the scenario is defined to be 3 degrees per 100 ft. After each decision point, the minimum curvature method\footnotemark{}\footnotetext{Detailed explanation of the minimum curvature method can be found in \citet{Kullawan2014-2}.} is used to calculate the well location for the next 32 discrete points based on the chosen alternative. 

At the beginning of the geosteering scenario, the well is positioned between 200 and 250 ft above the reservoir boundaries to accommodate a landing phase. The initial inclination is set to be 110 degrees. Throughout the landing phase, the steering alternatives remain unchanged. However, the constraint that bounded the inclination of the well is relaxed to be between 70 and 110 degrees during this phase. The landing phase concludes after the third decision point, equivalent to 960 ft after the initial point.

\begin{figure}[ht]
    \centering
    \includegraphics[scale=0.65]{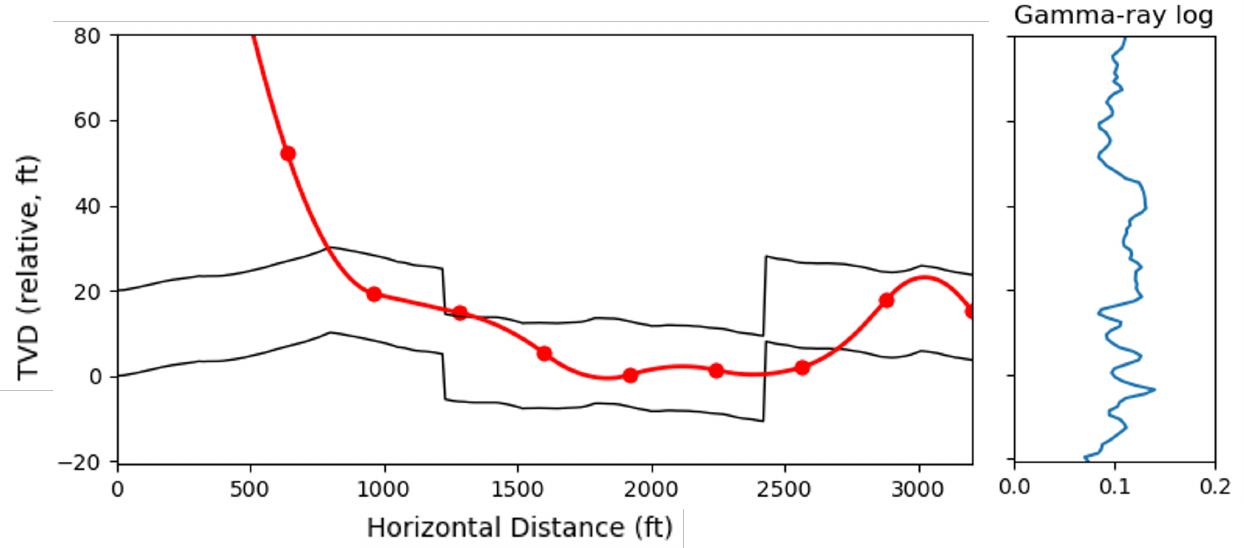}
    \caption{Illustration of the geosteering decision-making scenario. The trajectory of the drilled well is shown as a dotted red line, with each point indicating a decision point. Reservoir boundaries are randomly generated using functions from Alyaev (2022) and shown by black lines. The separate blue line plot shows the gamma-ray log data.} 
    \label{fig:strati}
\end{figure}

The geosteering decision-making scenario is shown in Figure \ref{fig:strati}. The vertical axis represents the TVD relative to the bottom reservoir boundary (y = 0). The trajectory of the drilled well is shown as a dotted red line, with each marker along the line representing a decision point. The black lines show the reservoir boundaries generated using functions from \citet{randomstrat}. A blue line in a separate plot shows the gamma-ray log data. This data comprises a portion of data shown in Figure \ref{fig:log}, representing the gamma-ray log from a specific depth interval. The scenario includes two faults at about 1250 ft and 2500 ft.

The objective of this scenario is to maximize reservoir contact, which means maximizing the length of the well placed within the reservoir. To achieve this objective, we define a reward function, $r$, which returns a value of $0$ if the well is positioned inside the reservoir and $-1$ if the well is outside the reservoir. The reward is accumulated over 32 discrete points and is provided to the decision-making agent after each decision point. Unlike the reward function used by the rule-based decision-making method, this reward function is quite straightforward. It does not penalize the decision-making agent based on the distance between the well and the reservoir boundaries. Instead, it focuses solely on whether the well is placed within the reservoir or not.

\begin{align}
    \label{eq:reward_sce}
    r= \begin{cases}
        0, & \text{if } \text{well} \text{ is inside}, \\
        -1, & \text{if } \text{well} \text{ is outside}.
    \end{cases}
\end{align}

\subsection{RL Implementation and Training Procedures}
Recall that we need to initialize the DNN before implementing the RL-based geosteering method. This initialization involves defining the state representation of the decision-making environment as the input to the DNN and the number of alternatives as the output of the DNN. Additionally, if the reward function has not been previously defined, we must define one that aligns with the overall objective of the scenario. 

As mentioned previously, the RL-Log and RL-Estimation methods have different state representations, resulting in distinct inputs to the DNN. Both methods, however, share similarities. First, they share a similarity in that both methods receive information about the latest inclination degree of the well and the current discrete point (or the distance from the initial point). Secondly, RL-Log and RL-Estimation use the same number of alternatives, resulting in 11 outputs, and the same reward function, $r$, as described in Section \ref{scenario}.

The state representation for the RL-Log method is the direct well-log data measured using the LWD tools. The decision-making agent uses this well-log data as its primary criterion for geosteering decisions. In contrast, the RL-Estimation method uses the outputs from PF as its state representation. The well-log data is processed to estimate the location of the well relative to the reservoir boundaries. The decision-making agent receives this estimated distance to the reservoir boundaries as its state representation. Additionally, the agent receives the corresponding estimates' probability relative to the other particles.

PF can be configured to generate multiple estimates. This aligns with the geosteering decision-making scenario, which is characterized by the possibility of having several solution modes or multi-modality, as reported by \citet{alyaev2022}. Therefore, obtaining several likely estimates and their corresponding probabilities is desirable. Given this, we will use two sets of PF estimates\footnotemark{}\footnotetext{The rule-based decision-making method will also use two sets of PF estimates.} as the state representation: one deterministic PF estimate (1 PF), representing the most likely solution of PF (particle with the highest weight), and five probabilistic PF estimates\footnotemark{}\footnotetext{The rationale behind using 5 PF estimates will be described in Section \ref{sec:PFresults}.} (5 PF). Note that the 5 PF state representation includes the deterministic and 4 probabilistic estimates.

After initializing the input and output of the DNN, we define additional details for the training procedure. For this study, we use PyTorch \citep{pytorch} to design the DNN and handle the training procedure. The default DNN configuration chosen is a three-hidden-layer network, each being a fully-connected linear layer and using a ReLU activation function. The number of neurons in the first and third hidden layers is set to two times the input size, while the second layer is set to have a number of neurons four times that of the input. For the RL-Log method, we adjust the configuration by increasing the number of neurons to twice the original amount. This results in 4-8-4 times the input size. This modification is implemented to improve the performance of RL-Log, which could still maintain relatively low computational requirements.

We also need to define a hyperparameter setting for the training procedure. The definition and value for each hyperparameter\footnotemark{}\footnotetext{More detailed information about the usage of each hyperparameter can be found in \citet{Mnih2015-mf}.} are summarized in Table \ref{tab:hyper}. The training procedure can be initiated with all the necessary components defined. The RL-Log and RL-Estimation will be trained using 11 training seeds drawn from a uniform distribution to ensure an unbiased comparison with the rule-based decision-making method. The evaluation procedure for all the trained decision-making agents from RL-Log and RL-Estimation methods, as well as the rule-based decision-making method, are described in the subsequent subsection.

\begin{table}[ht]
\caption{Hyperparameter setting used for the RL-Log and RL-Estimation methods}
\label{tab:hyper}
\resizebox{\textwidth}{!}{%
\begin{tabular}{|c|r|l|}
\hline
\multicolumn{1}{|c|}{\textbf{Hyperparameter}}                             & \multicolumn{1}{c|}{\textbf{Value}}   & \multicolumn{1}{c|}{\textbf{Definition}}                                                                                                            \\ \hline \hline
Discount factor ($\gamma$)                                                & 0.95                                  & \begin{tabular}[c]{@{}l@{}}Balances the importance of immediate and\\ future rewards during the training procedure.\end{tabular}                      \\ \hline
Learning rate ($\alpha$)                                                  & 0.0005                                & \begin{tabular}[c]{@{}l@{}}Controls the magnitude to which new information \\ overrides existing Q-values during the training procedure.\end{tabular} \\ \hline
Initial exploration                                                       & 1.0                                   & \begin{tabular}[c]{@{}l@{}}Initial probability of the agent taking a random \\ action instead of following the learned Q-values\end{tabular}        \\ \hline
Final exploration                                                         & 0.1                                   & \begin{tabular}[c]{@{}l@{}}Final probability of the agent taking a random \\ action instead of following the learned Q-values\end{tabular}          \\ \hline
\begin{tabular}[c]{@{}l@{}}Exploration \\ multiplier\end{tabular}         & 0.9997                                & \begin{tabular}[c]{@{}l@{}}Multiplier to decrease the value of exploration \\ parameter each episode\end{tabular}                                                                            \\ \hline
Minibatch size                                                            & 64                                    & \begin{tabular}[c]{@{}l@{}}Number of samples used in each iteration of the \\ training procedure to update the model parameters.\end{tabular}         \\ \hline
Memory size                                                               & 50000                                 & Capacity of the experience replay buffer                                                                                                            \\ \hline
\begin{tabular}[c]{@{}l@{}}Target network\\ update frequency\end{tabular} & 1000                                  & \begin{tabular}[c]{@{}l@{}}Controls the frequency to update the target \\ Q-value with the weights from the main Q-value\end{tabular}               \\ \hline
Training episodes                                                         & 20000                                 & Number of episodes to train the agent
                                    \\ \hline
\end{tabular}%
}
\end{table}

\subsection{Evaluation Procedure}\label{eval}
For the evaluation procedure, we generate 1000 reservoir boundary realizations using the same function used during the training procedure. However, there is a slight difference in the evaluation procedure between the rule-based decision-making method and the methods using RL. The evaluation for the rule-based decision-making strategy is straightforward, involving the calculation of the average reward obtained from the 1000 realizations. However, it is important to note that the average reward is calculated based on Equation \ref{eq:reward_sce} rather than Equation \ref{eq:reward_naive} to ensure a consistent comparison.

Since both RL-Log and RL-Estimation methods each train multiple decision-making agents, we need to calculate the median\footnotemark{}\footnotetext{Median is chosen over mean as the median of an odd dataset provides a more precise reward value from a trained decision-making agent.} of the average rewards obtained by all the agents. This is done using the following equations:
\begin{equation}\label{eq:median}
r_{RL} = \mathrm{Median}\left(\bar{r}_1,\bar{r}_2,\ldots,\bar{r}_{m}\right)
\end{equation}
\begin{equation}\label{eq:mean}
\bar{r} = \frac{\sum_{i=1}^{k} r_{i}}{k}
\end{equation}
Here, $m$ equals 11 (the number of training seeds), and $\bar{r}$ represents the average reward out of $k = 1000$ realizations. 

\section{Results and Discussions}
In this section, we present the results of the study along with comprehensive discussions about their significance. We begin by comparing the outputs from PF with the true reservoir boundaries. Subsequently, we show the training results obtained from the RL-Log and RL-Estimation methods. Finally, we conclude this section by presenting the evaluation results of all proposed methods.

\subsection{PF Estimates}\label{sec:PFresults}
\begin{figure}[ht]
    \centering
    \includegraphics[width=\textwidth]{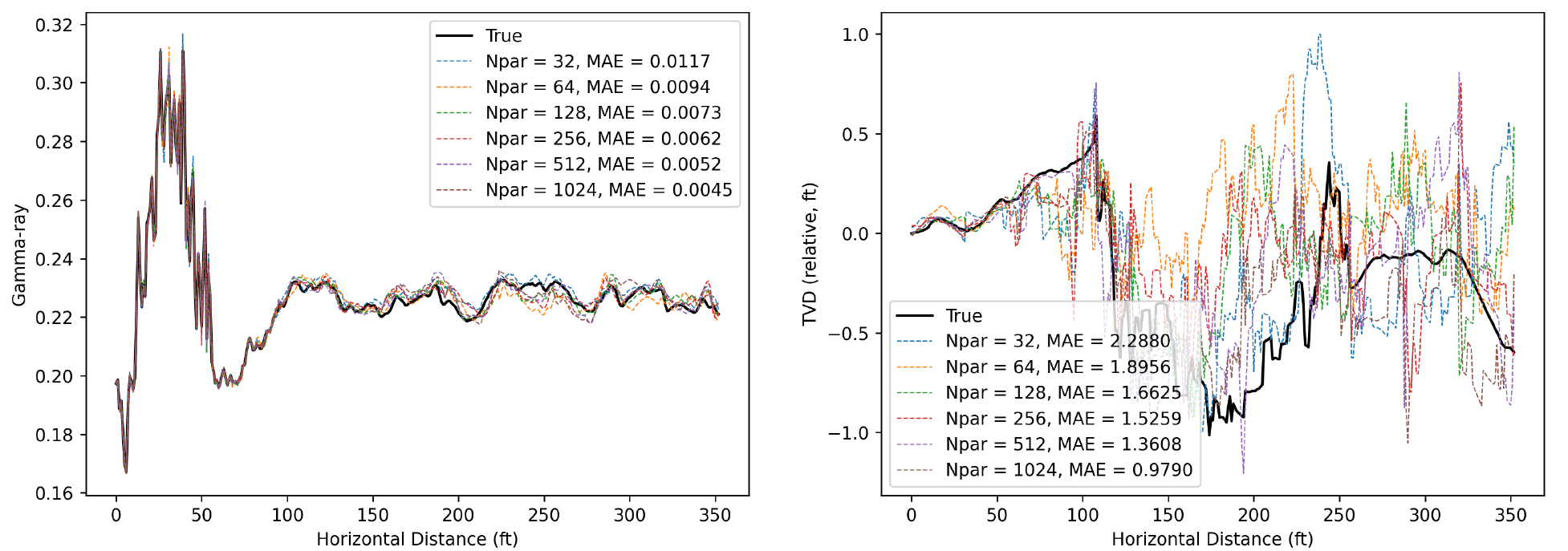}
    \caption{PF results across different numbers of particles, $N_\text{par}$, for two parameters: Gamma-ray (left) and Reservoir boundaries (right). The solid black line represents the true data. Dashed lines of varying colors correspond to the results obtained from different particle counts, with the MAE values provided in the legend. As evident, increasing the number of particles generally leads to a reduction in MAE, thereby improving the accuracy of the simulations.} 
    \label{fig:tempFig1}
\end{figure}

The estimates for gamma-ray and its corresponding reservoir boundaries for PFs starting with $N_\text{par} = 32$ and up to $N_\text{par} = 1024$ along with the corresponding MAE values are shown in Figure \ref{fig:tempFig1}. Each line in the figure represents the best particle (the particle with the highest weight) for each $N_\text{par}$. The figure is generated based on 100 different reservoir realizations.

One of the objectives of our study with PF is to determine the optimal number of particles before integrating it with RL. We established a Mean Absolute Error (MAE) threshold of 0.01 to guide this selection. Observing the MAE of the best particle, as presented in the left panel in Figure \ref{fig:tempFig1}, it is evident that starting from 64 particles, the MAE values consistently fall below the 0.01 mark. Nevertheless, while increasing the number of particles enhances accuracy, it also increases the computational requirements associated with PF-based estimates. Though the 64-particle configuration yielded an MAE of 0.0094, we opted for 128 particles, ensuring a more conservative MAE value that lies well beneath our set threshold, thus providing a balance between precision and computational efficiency.

A distinct observation can be drawn regarding the performance of PF in estimating gamma-ray and reservoir boundaries. The left panel in Figure \ref{fig:tempFig1} shows that the Particle Filter is adept at selecting the best particle that matches with the true gamma-ray values. However, a contrasting scenario is shown in the right panel, where the corresponding reservoir boundaries are incongruous with the true values. This observation indicates that there could be multiple possible realizations of the reservoir boundaries that result in similar gamma-ray observations and emphasizes the need to consider multiple particles to deal with the ill-posed inverse problem. 

\begin{figure}[ht]
    \centering
    \includegraphics[width=\textwidth]{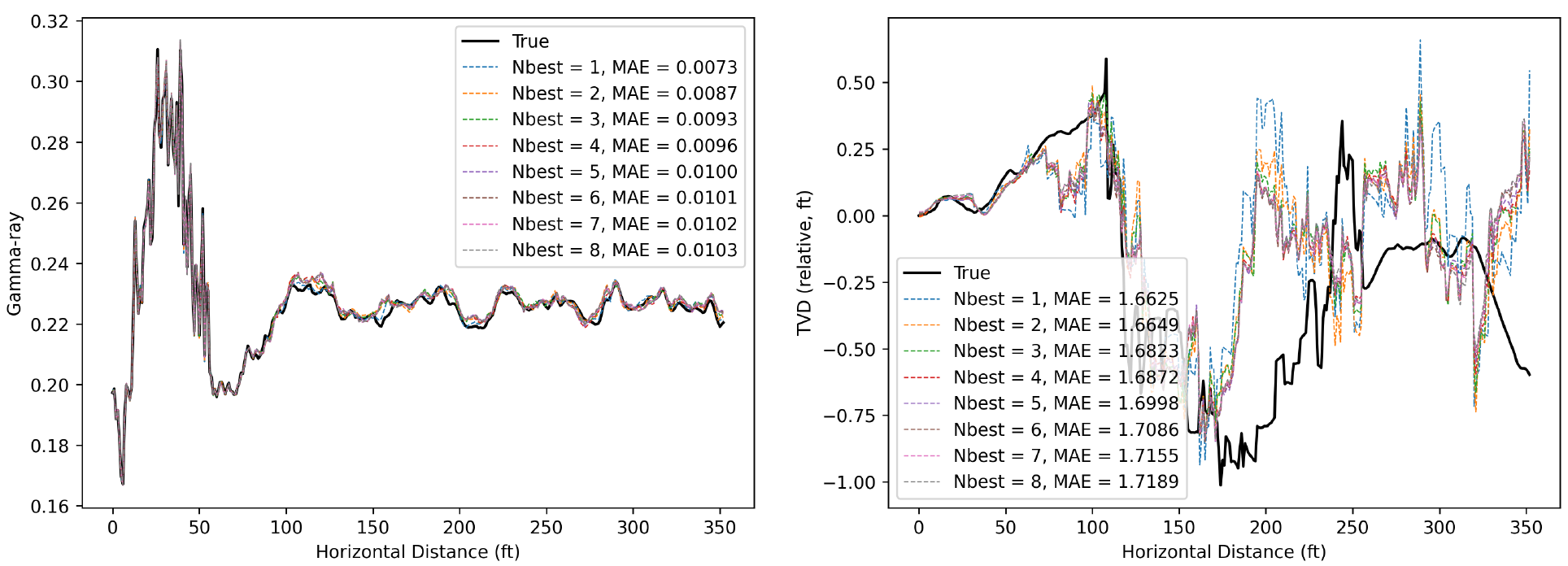}
    \caption{PF results across different values of number of best particles, $N_\text{best}$, for two parameters: Gamma-ray (left) and Reservoir boundaries (right). The solid black line represents the true data. Dashed lines of varying colors correspond to the results obtained from different $N_\text{best}$ values, with the Mean MAE values provided in the legend. As evident, increasing the $N_\text{best}$ values shows a slight increase in MAE, indicating a deviation from the true values.} 
    \label{fig:tempFig2}
\end{figure}

Figure \ref{fig:tempFig2} highlights the relationship between the number of best particles, $N_\text{best}$, and the corresponding MAE for $N_\text{par}$. Each line represents the mean of $N_\text{best}$ particles, ranging from 1 to 8 particles. As we increase the number of particles from 1 to 8, the MAE is increasing. Note that the MAE for $N_\text{best} = 1$ aligns precisely with the MAE observed for $N_\text{par} = 128$ particles from Figure \ref{fig:tempFig1}.  Of the eight particles shown, the $N_\text{best} = 5$ exhibits an MAE that aligns precisely with our predefined threshold of 0.01. The right panel shows that the choice between using the best-performing single particle ($N_\text{best} = 1$) or employing multiple particles has minimal impact on the mean absolute error (MAE) of the true reservoir boundaries. Therefore, we opt to use the probabilistic solution of $N_\text{best} =$ 5 particles, including the most likely solution (the best particle) for the integration with RL.


\subsection{Training Results}

\begin{figure}[ht]
    \centering
    \includegraphics[scale=0.4]{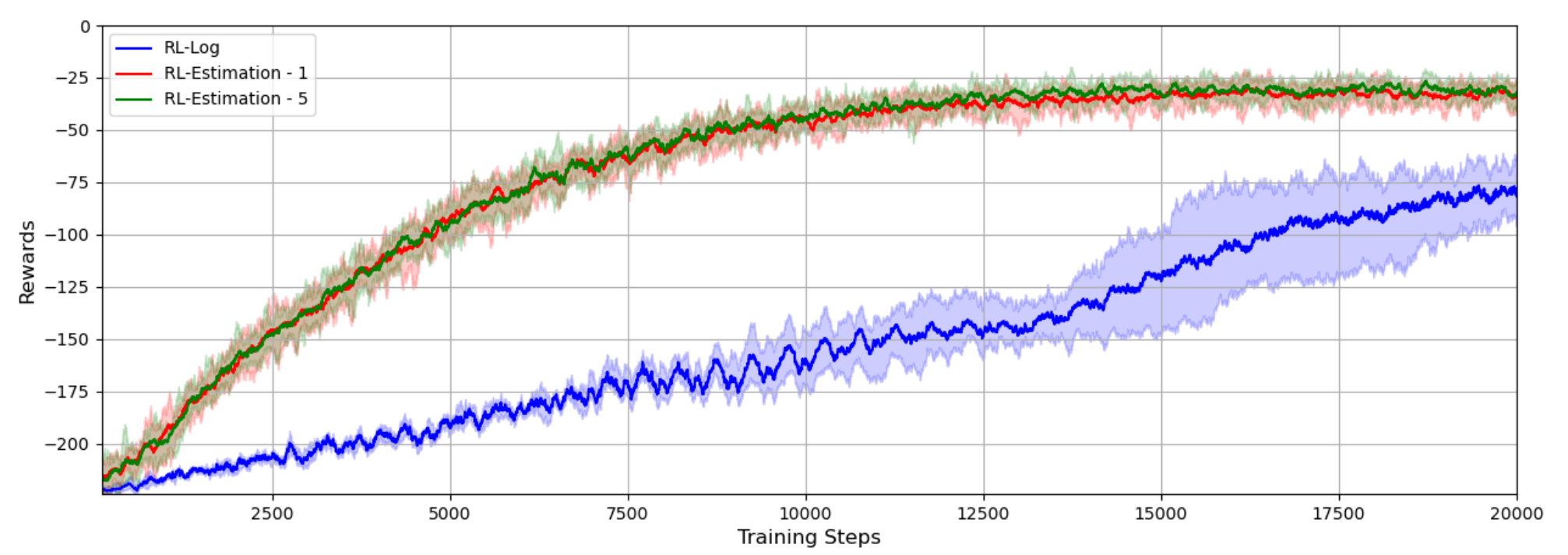}
    \caption{The graph shows average training rewards with shaded areas indicating confidence intervals. Average rewards over the last 100 training episodes are used for a smoother depiction of training progress and to reduce short-term fluctuations.}. 
    \label{fig:trainres}
\end{figure}

\textbf{Results. }Figure \ref{fig:trainres} shows the training results of the RL-Log method and the two RL-Estimation methods from 11 different training seeds. The lines in the graph represent the average values, while the shaded areas depict the corresponding confidence intervals. The legend provides the key to interpreting the methods displayed in the figure. We calculate the average rewards acquired during the last 100 training episodes to construct this figure. This approach ensures a smoothed representation of the training procedure, effectively mitigating the impact of short-term fluctuations.

All methods gradually improve the overall rewards throughout the training, as seen by the overall increase in the rewards. Initially, all methods start with a reward of approximately -220, corresponding to 2 percent reservoir contact. As training progresses, the RL-Log method achieves an average reward of approximately -75, while both RL-Estimation methods achieve an average reward of approximately -30. Furthermore, the RL-Estimation methods demonstrate a more stable training procedure than the RL-Log method, as seen by their relatively smaller shaded areas towards the end of the training procedure. 

Another important aspect is the computational requirements of training a single seed. While the RL-Log method achieves lower results at the end of the training procedure, it only requires approximately 30 minutes for training. In contrast, the RL-Estimation methods have expensive computational requirements. The RL-Estimation method using 1 PF input requires approximately 300 minutes for training. On the other hand, the RL-Estimation method using 5 PF inputs requires approximately 420 minutes for training. 

\textbf{Discussions. }With a training time of only 30 minutes, the RL-Log method can be considered a computationally efficient method. However, this advantage comes at the expense of possibly yielding relatively worse decision-making strategies than the RL-Estimation methods. On the other hand, despite requiring more computational requirements, the RL-Estimation methods show superior performance and stability during training. The stability suggests that the RL-Estimation methods are less affected by random training seeds than the RL-Log method. 

The training results of the RL-Estimation method using 1 and 5 PF inputs appeared similar. A more detailed look at the graph suggests a slight advantage for the RL-Estimation method using 5 PF inputs, particularly in the later stages of the training procedure. However, the difference in training results can be seen as insignificant, given the notable difference in their computational requirements. The difference in computational requirements between 1 and 5 PF inputs is primarily attributed to the number of neurons within the DNN. As the input's dimensionality increases, there is a corresponding increase in the computational requirements for the training procedure.

On the other hand, the considerable difference in computational requirements between the RL-Log and RL-Estimation methods is primarily attributed to the use of PF at each discrete point. A closer look at a single episode in the RL-Estimation method reveals that PF accounts for approximately 95 percent of the computational requirements. The percentage is notably high and opens the possibility of exploring other methods, such as DNN-based state estimation methods, to reduce computational requirements while maintaining the accuracy of estimates. 

Nevertheless, based on the overall training results, the RL-Estimation method, using either 1 or 5 PF inputs, appears to be the preferred method for solving realistic geosteering decision-making scenarios. This holds especially true in cases where sufficient computational resources are available. To further validate the training results, we compare all proposed methods more comprehensively based on the procedure described in Section \ref{eval} in the following subsection.

\subsection{Evaluation Results}

\textbf{Results. }Table \ref{tab:results} shows the methods evaluated in the study, their corresponding input parameters, rewards yielded, reservoir contact percentages, and the MAE of input values. To ensure consistent comparisons, the rewards are calculated after the landing phase, thus excluding the unavoidable -1 rewards. For the rule-based and RL-Estimation methods, results are provided for both using the 1 and 5 PF inputs. 

\begin{table}[ht]
\caption{Evaluation results for the Rule-based decision-making, RL-Log, and RL-Estimation methods}
\centering
\label{tab:results}
\begin{tabular}{|c|c|r|r|r|}
\hline
\textbf{Methods}       & \textbf{Input}              & \multicolumn{1}{c|}{\textbf{Rewards}} & \multicolumn{1}{c|}{\begin{tabular}[c]{@{}c@{}}\textbf{Reservoir} \\ \textbf{contact (\%)}\end{tabular}} & \multicolumn{1}{c|}{\begin{tabular}[c]{@{}c@{}}\textbf{MAE of} \\ \textbf{Input}\end{tabular}}\\ \hline\hline
Rule-based\footnotemark{}         & 1 Look-ahead   &  -101.36                            & 54.75   &   4.27                                 \\ \hline 
Rule-based\footnotemark[\value{footnote}]         & 5 Look-ahead & -105.95                             & 52.70  &   4.27                                  \\ \hline
RL-Log        & Gamma-ray          & -74.27                             & 66.84    &   -                                \\ \hline
RL-Estimation & 1 PF   & -25.36                            & 88.67    & 1.29                                  \\ \hline
RL-Estimation & 5 PF & -24.99                             & 88.91    & 1.31\\ \hline
\end{tabular}%
\end{table}

The application of the rule-based decision-making method results in rewards of -101.36 and -105.95 for inputs of 1 and 5 look-ahead estimates, respectively. These rewards correspond to reservoir contacts of 54.75 and 52.70 percent, respectively. The RL-Log method, which utilizes RL with gamma-ray log data as input, yields a reward of -74.27 and a reservoir contact of 66.84 percent. The RL-Estimation methods perform significantly better than the other methods, achieving rewards of -25.36 and -24.99 for 1 and 5 PF inputs, respectively, corresponding to reservoir contact of 88.68 and 88.85 percent.
\footnotetext{The landing phase of the rule-based decision-making method is assisted by a trained RL method, as the former is incapable of optimizing decisions during this phase.}

The MAE calculation varies with the method used as the decision-making criterion. For the rule-based decision-making method, the MAE value refers to the difference between the true reservoir boundaries and the random look-ahead estimates, resulting in a value of 4.27. The RL-Estimation method uses different calculations where the MAE value refers to the difference between the true reservoir boundaries and the outputs yielded from PF. This results in a lower MAE value of 1.29 and 1.31 for the methods using 1 and 5 PF inputs, respectively. Contrary to the other two methods, the RL-Log method is not subjected to the MAE calculation, given the underlying assumption of accurate gamma-ray log data. 

\begin{figure}[ht]
    \centering
    \includegraphics[scale=0.29]{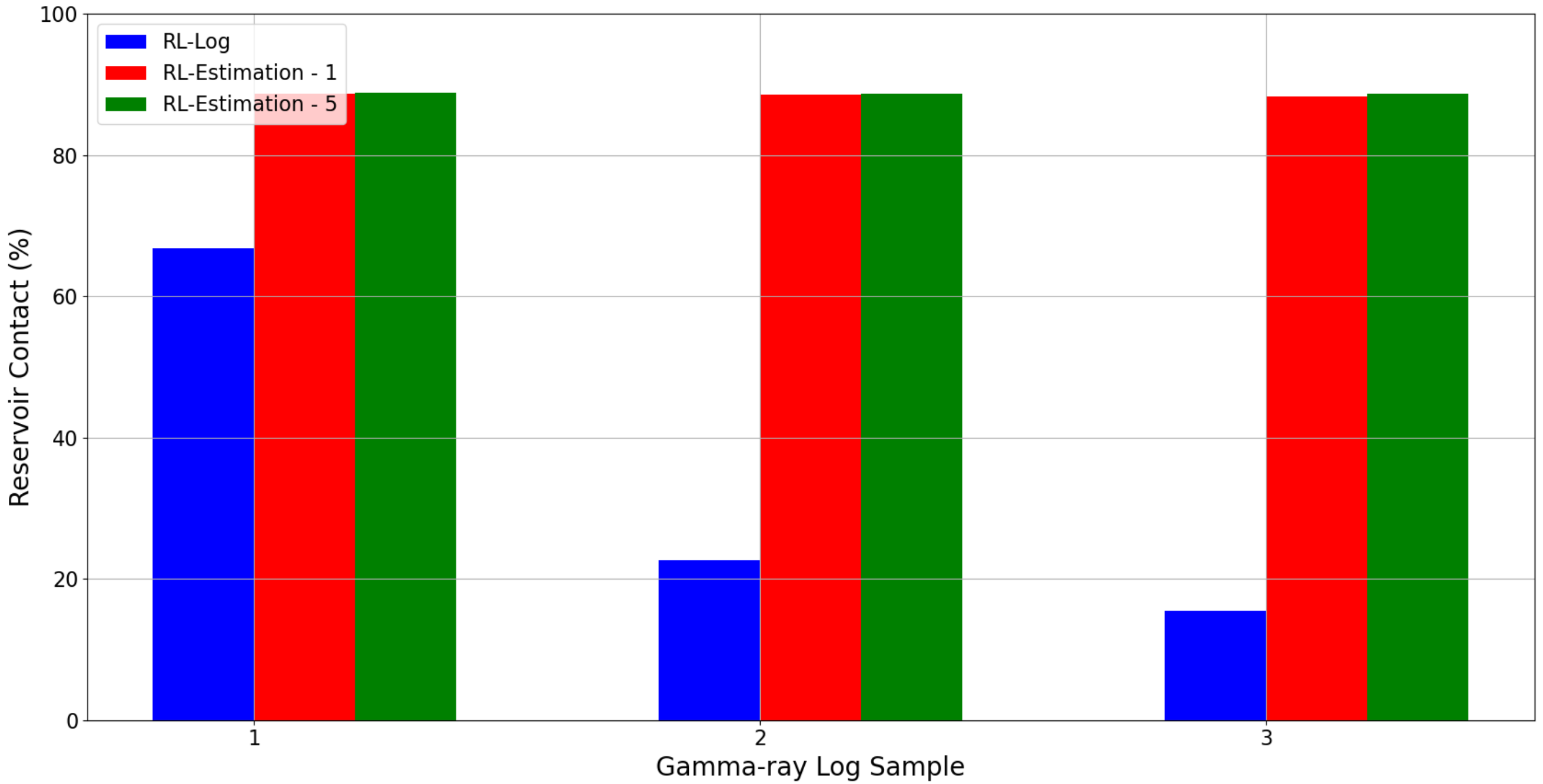}
    \caption{Evaluation results of the RL-Log and RL-Estimation methods, considering different TVD at which the gamma-ray log data is sampled.}. 
    \label{fig:logestimate}
\end{figure}

Figure \ref{fig:logestimate} shows the results of evaluating the RL-Log and RL-Estimation methods on various gamma-ray log samples. We have excluded the rule-based decision-making method from this evaluation due to its inferior overall performance. The gamma-ray log samples correspond to different TVDs in Figure \ref{fig:log}, which indicates the assumed reservoir depth. Specifically, Sample 1 in the figure refers to the gamma-ray log data used during the training and evaluation procedure described above. It serves as the baseline against which the other two samples will be compared.

The figure shows a significant decline in the reservoir contact of the RL-Log method. This decline is observed when the method is evaluated using different gamma-ray log data compared to the data used in its training. The reservoir contact results for two different gamma-ray log samples deviate significantly from the baseline of 66.84 percent. On the other hand, both RL-Estimation methods show a consistent pattern across all samples, maintaining reservoir contacts identical to their respective baseline of approximately 88 percent. Furthermore, the figure shows that the RL-Estimation method using 5 PF inputs consistently outperforms the method using 1 PF input, despite the small difference between them.

\begin{figure}[ht]
    \centering
    \includegraphics[scale=0.29]{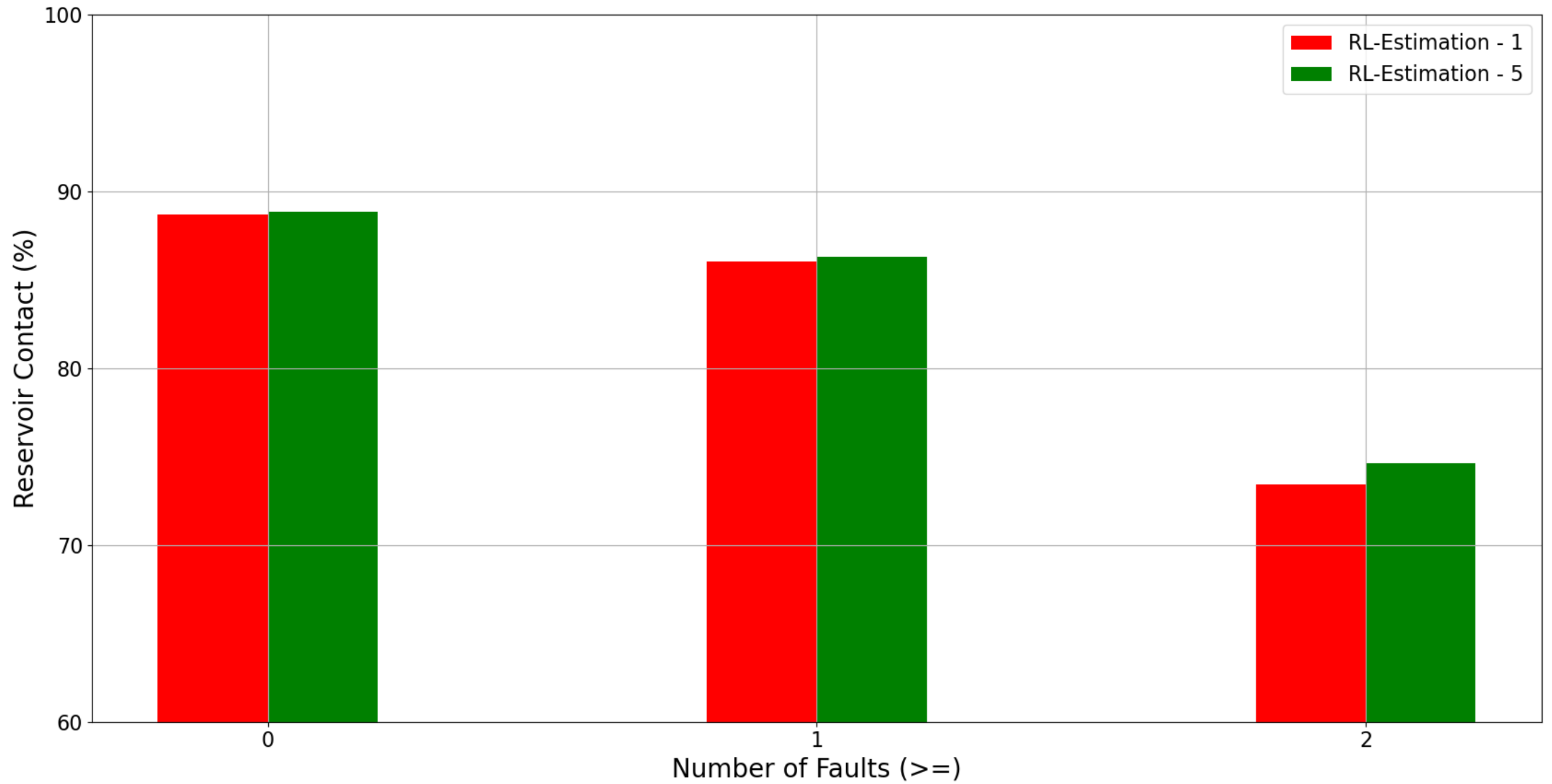}
    \caption{Evaluation results of the RL-Estimation methods, considering different minimum numbers of encountered faults.}. 
    \label{fig:estimatefault}
\end{figure}

Figure \ref{fig:estimatefault} shows the results of evaluating the RL-Estimation methods, focusing on the minimum number of encountered faults. We have excluded the RL-Log method from this evaluation due to its inferior performance in the prior evaluation. We use three different thresholds for the minimum number of faults. Reservoir contact for faults $\geq 0$ refers to the percentage obtained from the previous evaluation. The figure shows an expected decrease in reservoir contact as the minimum fault number increases. As a result, when faults $\geq 2$, the reservoir contact reaches its minimum value at approximately 74 percent for both the RL-Estimation using 1 and 5 PF inputs. Furthermore, the figure shows a consistent trend observed in the previous evaluation, in which the RL-Estimation method using 5 PF inputs consistently outperforms the one using 1 PF input. This difference becomes slightly more noticeable in scenarios involving more faults.

\textbf{Discussions. }The rule-based decision-making method performs the worst. This is attributed to two key factors: adopting the greedy optimization method and using the na\"{i}ve estimation method for generating look-ahead estimates. Selecting locally optimized decisions is suitable for simpler sequential decision-making scenarios. However, its application in geosteering decision-making scenarios is ineffective, especially compared to RL, which integrates a learning process and its impact on future decisions.

Additionally, the input provided to the rule-based method is also suboptimal. While the initial reference points for these estimates are derived from the best-performing particles from PF, the subsequent look-ahead estimates are generated in a stochastic manner. This explanation is validated by the MAE values presented in the table. Specifically, the rule-based decision-making methods optimize decisions using an input featuring an MAE of 4.27. Moreover, the performance of the rule-based method with 5 look-ahead estimates is slightly worse than that with only 1 look-ahead estimate. It can be attributed to the accumulation of high MAE values from all five estimates.

The RL-Log method improves rewards and reservoir contact percentages compared to the rule-based decision-making method. This improvement is primarily attributed to using RL for optimizing decisions. The RL-Log method yields improvement in results even when provided with gamma-ray log data, which we considered a less adaptable decision-making criterion than the distance to reservoir boundaries. However, the improvements observed in the RL-Log method are restricted to the gamma-ray log data used during its training. The resulting decision-making strategy is not applicable to different gamma-ray log samples. Consequently, attempting to apply it to such samples will lead to suboptimal decisions and low overall reservoir contact.

In contrast, the RL-Estimation methods yield consistent results across all gamma-ray log samples, whether using 1 or 5 PF inputs. The results demonstrate the ability of the RL-Estimation method to produce an optimal decision-making strategy irrespective of the gamma-ray log data. The results also confirm that using the distance to reservoir boundaries, represented here by PF inputs, is a more flexible decision-making criterion than direct well-log data. The significant difference in reservoir contact justifies the computational requirements associated with integrating PF into the framework of the RL-Estimation method. Considering the results from all methods, it is evident that the integration of RL and PF establishes a synergistic dependency between the two, leading to significantly enhanced outcomes in optimizing geosteering decisions.

Further evaluation of the RL-Estimation methods indicates that using 5 PF inputs results in consistently better performance than using 1 PF input across all evaluation parameters. The results demonstrate the additional value created from using probabilistic PF inputs compared to relying solely on deterministic inputs, especially in scenarios with a possibility of multi-modality occurrences. However, the insignificant improvement in results may indicate that there are only a few multi-modality occurrences in this scenario. 

Nevertheless, it is evident that both RL-Estimation methods encounter challenges from an increased number of faults. This result can be attributed to the lack of look-ahead estimates. However, PF is ill-suited for generating accurate look-ahead estimates, as demonstrated by the rule-based decision-making method. Therefore, an alternative state estimation method is needed to evaluate the value created from integrating look-ahead estimates within the RL framework.

\section{Benchmarking Study}
In this benchmarking study, we use an RL that uses the distance to the true reservoir boundaries as its decision-making criterion as a benchmark for the RL-Estimation method, thus eliminating the errors arising from the state estimation method. The distance to the true reservoir boundaries represents the theoretically optimal criterion for RL, assuming the real condition of not having access to the look-ahead reservoir boundaries. This benchmarking study evaluates whether the outputs derived from PF are sufficient as the decision-making criterion for the RL-Estimation method or if a more accurate state estimation method is needed.

Furthermore, we provide an additional benchmark for the RL-Estimation method. Specifically, we compare the method against RL using the distance to the true and look-ahead reservoir boundaries as the decision-making criterion. It represents the theoretically optimal criterion for RL. Consequently, it is expected to yield the optimal results when using RL as the decision optimization method.

\begin{figure}[ht]
    \centering
    \includegraphics[scale=0.735]{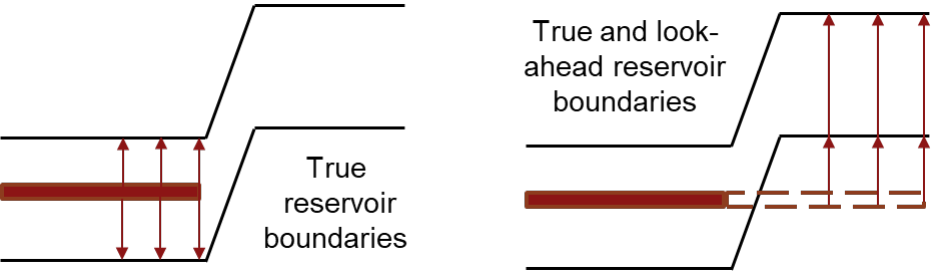}
    \caption{Illustration of the decision-making criterion, indicated by the red arrows, used for the benchmarking study. The black lines show the true reservoir boundaries, while the solid red lines show the trajectory of the drilled well. The figure on the left illustrates the distance to the true reservoir boundaries as the decision-making criterion for RL. The right-hand side figure illustrates the distance from the projected well, indicated by dashed red lines, to the true and look-ahead reservoir boundaries.} 
    \label{fig:benchmarksetup}
\end{figure}

Figure \ref{fig:benchmarksetup} illustrates the decision-making criterion provided to RL for the benchmarking study. Similar to the figures in Section \ref{proposed}, the black lines show the true reservoir boundaries, while the solid red lines show the trajectory of the drilled well. The dashed red lines show the projected well used to calculate the distance to the true and look-ahead reservoir boundaries. The red arrows show the decision-making criterion for each benchmarking study. Specifically, the left-hand side figure illustrates the distance to the true reservoir boundaries, while the right-hand side figure illustrates the distance to the true and look-ahead reservoir boundaries.

\begin{figure}[ht]
    \centering
    \includegraphics[scale=0.29]{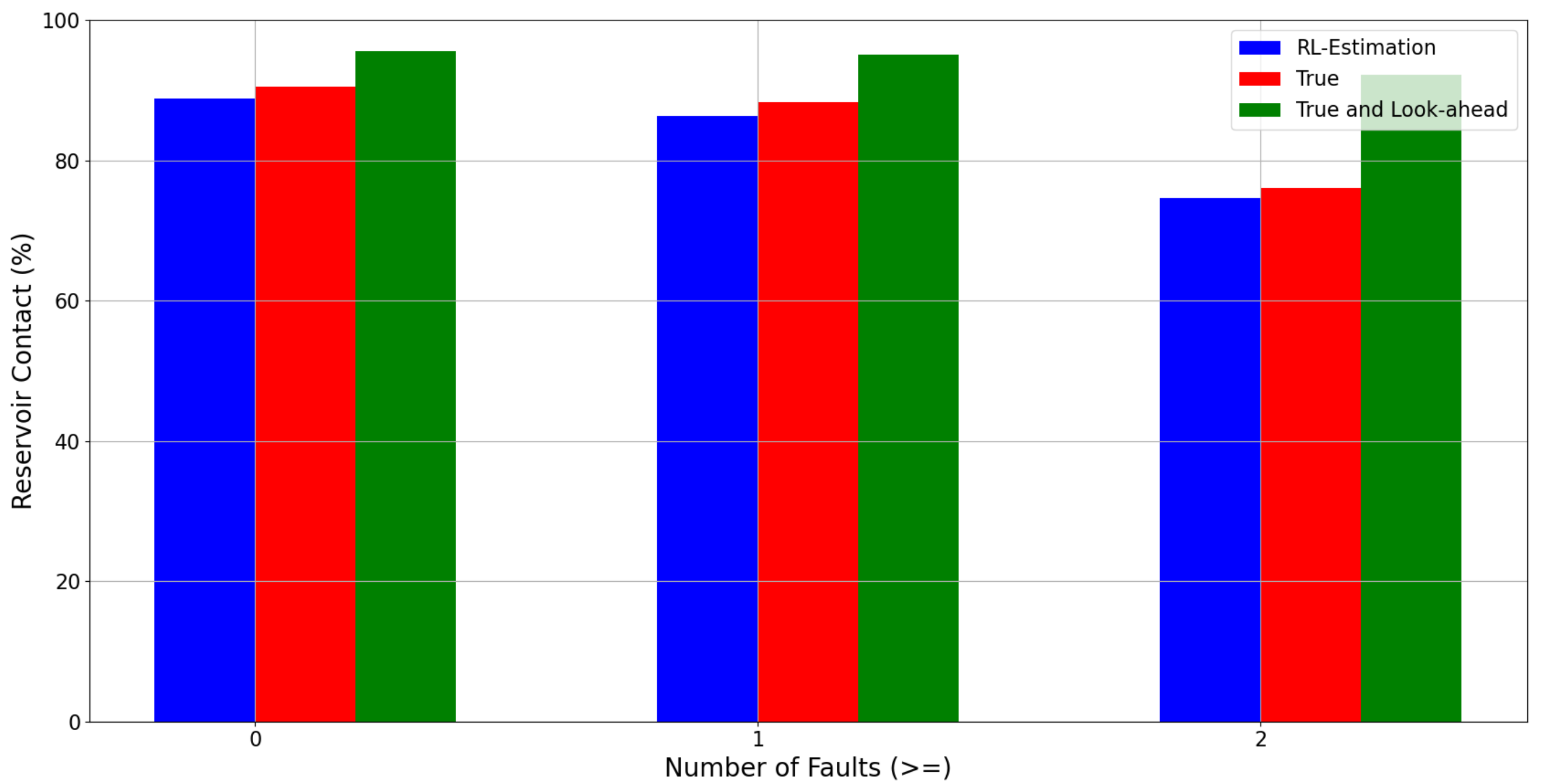}
    \caption{Benchmarking study of the RL-Estimation method to the theoretically possible results, considering different minimum numbers of encountered faults.}. 
    \label{fig:benchmark}
\end{figure}

\textbf{Results.} Figure \ref{fig:benchmark} shows the results of the benchmarking study. The evaluation is conducted across different numbers of encountered faults. The figure shows a consistent trend wherein RL using the distance to the true reservoir boundaries consistently yields 1 to 2 percent better reservoir contact compared to the RL-Estimation method. Furthermore, providing the distance to the true and look-ahead reservoir boundaries to RL leads to the optimal results, resulting in a consistent reservoir contact of above 90 percent. The most notable difference occurs in scenarios where faults $\geq 2$, a scenario in which the RL-Estimation method is known to struggle.

\textbf{Discussions.} The results of the benchmarking study show that the RL-Estimation method yields comparable results to RL provided by the distance to the true reservoir boundaries. With a difference in reservoir contact of only 1 to 2 percent, we can reasonably consider that PF yields sufficiently accurate outputs for the RL-Estimation method. The only drawback of PF is its expensive computational requirements, which take approximately 95 percent of the computational requirements for the RL-Estimation method. Despite this, the RL-Estimation method requires less than a second to evaluate a single geosteering decision-making scenario, demonstrating its efficiency in practical applications.

The second benchmarking study shows that providing RL with the distance to the true and look-ahead reservoir boundaries results in the highest reservoir contact. While these results are theoretical in nature, they show the potential of RL in further optimizing geosteering decisions that will lead to optimal results, particularly the ones involving multiple faults. However, it requires a state estimation method capable of providing accurate look-ahead estimates. If the level of accuracy of the method is similar to PF, it is possible to achieve a reservoir contact optimal reservoir contact comparable to the maximum benchmark.

\section{Conclusions}
In this study, we introduce "RL-Estimation", a novel geosteeering-decision method 
for optimal results in realistic geosteering decision-making scenarios. 
Building upon the verified decision-making capabilities of RL methods, we developed the RL-Estimation method to align with common practice in geosteering. The practice refers to an informed decision-making process driven by the interpretation of well-log data. To achieve this, the RL-Estimation method relies on PF, a state estimation method, to yield an estimated distance to the reservoir boundaries.

The RL-Estimation method performs better than the two alternative methods developed and discussed in this study. Specifically, these methods are the RL-Log method, which relies on RL without including PF, and the rule-based decision-making method, which solely uses PF without integration with RL. Upon evaluation, the RL-Log method suffers when evaluating data other than the one used during the training procedure. On the other hand, the rule-based decision-making method is the least effective method across the metrics considered. Therefore, it becomes evident that there is a dependency between RL and PF in yielding optimized geosteering decisions.

The RL-Estimation method achieves comparable results to the optimal benchmark, which indicates a high level of estimates accuracy provided by PF. However, the RL-Estimation method requires relatively high computational requirements as a result of using PF. Therefore, future studies should build upon this work by exploring more complex state estimation methods, such as the DNN-based method, as the alternative to PF. This has the potential to ensure comparable results to the RL-Estimation method while significantly reducing computational requirements. Furthermore, the benchmarking studies reveals the potential of exploring the additional value created by using look-ahead estimates as the decision-making criterion for RL, most notably for faulted geosteering scenarios.

\section*{Acknowledgements}
This work is part of the Center for Research-based Innovation DigiWells: Digital Well Center for Value Creation, Competitiveness and Minimum Environmental Footprint (NFR SFI project no. 309589, https://DigiWells.no). The center is a cooperation of NORCE Norwegian Research Centre, the University of Stavanger, the Norwegian University of Science and Technology (NTNU), and the University of Bergen. It is funded by Aker BP, ConocoPhillips, Equinor, TotalEnergies, Vår Energi, Wintershall Dea, and the Research Council of Norway.

\section*{Declaration of generative AI and AI-assisted technologies in the writing process}
During the preparation of this work, the authors used ChatGPT in order to improve the readability. After using this tool/service, the authors reviewed and edited the content as needed and take full responsibility for the content of the publication.

\bibliographystyle{elsarticle-harv} 
\bibliography{main}

\begin{thebibliography}{39}
\expandafter\ifx\csname natexlab\endcsname\relax\def\natexlab#1{#1}\fi
\providecommand{\url}[1]{\texttt{#1}}
\providecommand{\href}[2]{#2}
\providecommand{\path}[1]{#1}
\providecommand{\DOIprefix}{doi:}
\providecommand{\ArXivprefix}{arXiv:}
\providecommand{\URLprefix}{URL: }
\providecommand{\Pubmedprefix}{pmid:}
\providecommand{\doi}[1]{\href{http://dx.doi.org/#1}{\path{#1}}}
\providecommand{\Pubmed}[1]{\href{pmid:#1}{\path{#1}}}
\providecommand{\bibinfo}[2]{#2}
\ifx\xfnm\relax \def\xfnm[#1]{\unskip,\space#1}\fi
\bibitem[{Alyaev(2022)}]{randomstrat}
\bibinfo{author}{Alyaev, S.}, \bibinfo{year}{2022}.
\newblock \bibinfo{title}{{Supporting Data for: Direct multi-modal inversion of geophysical logs using deep learning}}.
\newblock \DOIprefix\doi{10.18710/1F9GYH}.
\bibitem[{Alyaev et~al.(2022)Alyaev, Ambrus, Jahani and Elsheikh}]{alyaev2022sequential}
\bibinfo{author}{Alyaev, S.}, \bibinfo{author}{Ambrus, A.}, \bibinfo{author}{Jahani, N.}, \bibinfo{author}{Elsheikh, A.H.}, \bibinfo{year}{2022}.
\newblock \bibinfo{title}{Sequential multi-realization probabilistic interpretation of well logs and geological prediction by a deep-learning method}, in: \bibinfo{booktitle}{SPWLA Annual Logging Symposium}, \bibinfo{organization}{SPWLA}. p. \bibinfo{pages}{D051S022R001}.
\bibitem[{Alyaev and Elsheikh(2022)}]{alyaev2022}
\bibinfo{author}{Alyaev, S.}, \bibinfo{author}{Elsheikh, A.H.}, \bibinfo{year}{2022}.
\newblock \bibinfo{title}{Direct multi-modal inversion of geophysical logs using deep learning}.
\newblock \bibinfo{journal}{Earth and Space Science} \bibinfo{volume}{9}.
\newblock \DOIprefix\doi{10.1029/2021EA002186}.
\bibitem[{Alyaev et~al.(2018)Alyaev, Hong and Bratvold}]{Alyaev2018}
\bibinfo{author}{Alyaev, S.}, \bibinfo{author}{Hong, A.}, \bibinfo{author}{Bratvold, R.B.}, \bibinfo{year}{2018}.
\newblock \bibinfo{title}{Are you myopic, naïve or farsighted about your geosteering decisions?}
\newblock \bibinfo{journal}{Second EAGE/SPE Geosteering and Well Placement Workshop} \bibinfo{volume}{2018}, \bibinfo{pages}{1--5}.
\newblock \DOIprefix\doi{https://doi.org/10.3997/2214-4609.201803217}.
\bibitem[{Alyaev et~al.(2021)Alyaev, Ivanova, Holsaeter, Bratvold and Bendiksen}]{ALYAEV2021}
\bibinfo{author}{Alyaev, S.}, \bibinfo{author}{Ivanova, S.}, \bibinfo{author}{Holsaeter, A.}, \bibinfo{author}{Bratvold, R.B.}, \bibinfo{author}{Bendiksen, M.}, \bibinfo{year}{2021}.
\newblock \bibinfo{title}{An interactive sequential-decision benchmark from geosteering}.
\newblock \bibinfo{journal}{Applied Computing and Geosciences} \bibinfo{volume}{12}, \bibinfo{pages}{100072}.
\newblock \DOIprefix\doi{10.1016/j.acags.2021.100072}.
\bibitem[{Alyaev et~al.(2019)Alyaev, Suter, Bratvold, Hong, Luo and Fossum}]{ALYAEV2019}
\bibinfo{author}{Alyaev, S.}, \bibinfo{author}{Suter, E.}, \bibinfo{author}{Bratvold, R.B.}, \bibinfo{author}{Hong, A.}, \bibinfo{author}{Luo, X.}, \bibinfo{author}{Fossum, K.}, \bibinfo{year}{2019}.
\newblock \bibinfo{title}{A decision support system for multi-target geosteering}.
\newblock \bibinfo{journal}{Journal of Petroleum Science and Engineering} \bibinfo{volume}{183}, \bibinfo{pages}{106381}.
\newblock \DOIprefix\doi{10.1016/j.petrol.2019.106381}.
\bibitem[{Bakr et~al.(2017)Bakr, Pardo and Torres-Verd{\'\i}n}]{bakr2017fast}
\bibinfo{author}{Bakr, S.A.}, \bibinfo{author}{Pardo, D.}, \bibinfo{author}{Torres-Verd{\'\i}n, C.}, \bibinfo{year}{2017}.
\newblock \bibinfo{title}{Fast inversion of logging-while-drilling resistivity measurements acquired in multiple wells}.
\newblock \bibinfo{journal}{Geophysics} \bibinfo{volume}{82}, \bibinfo{pages}{E111--E120}.
\bibitem[{Bannister(2017)}]{varDAReview}
\bibinfo{author}{Bannister, R.N.}, \bibinfo{year}{2017}.
\newblock \bibinfo{title}{A review of operational methods of variational and ensemble-variational data assimilation}.
\newblock \bibinfo{journal}{Quarterly Journal of the Royal Meteorological Society} \bibinfo{volume}{143}, \bibinfo{pages}{607--633}.
\newblock \DOIprefix\doi{https://doi.org/10.1002/qj.2982}.
\bibitem[{Bratvold and Begg(2010)}]{bratvoldbook}
\bibinfo{author}{Bratvold, R.B.}, \bibinfo{author}{Begg, S.H.}, \bibinfo{year}{2010}.
\newblock \bibinfo{title}{{Making Good Decisions}}.
\newblock \bibinfo{publisher}{Society of Petroleum Engineers}.
\newblock \DOIprefix\doi{10.2118/9781555632588}.
\bibitem[{Chen et~al.(2014)Chen, Lorentzen and Vefring}]{Chen2014}
\bibinfo{author}{Chen, Y.}, \bibinfo{author}{Lorentzen, R.J.}, \bibinfo{author}{Vefring, E.H.}, \bibinfo{year}{2014}.
\newblock \bibinfo{title}{{Optimization of Well Trajectory Under Uncertainty for Proactive Geosteering}}.
\newblock \bibinfo{journal}{SPE Journal} \bibinfo{volume}{20}, \bibinfo{pages}{368--383}.
\newblock \DOIprefix\doi{10.2118/172497-PA}.
\bibitem[{Cheraghi et~al.(2022)Cheraghi, Alyaev, Hong, Kuvaev, Clark, Zhuravlev and Bratvold}]{yasaman}
\bibinfo{author}{Cheraghi, Y.}, \bibinfo{author}{Alyaev, S.}, \bibinfo{author}{Hong, A.}, \bibinfo{author}{Kuvaev, I.}, \bibinfo{author}{Clark, S.}, \bibinfo{author}{Zhuravlev, A.}, \bibinfo{author}{Bratvold, R.B.}, \bibinfo{year}{2022}.
\newblock \bibinfo{title}{What can we learn after 10,000 geosteering decisions?}, in: \bibinfo{booktitle}{SPE/AAPG/SEG Unconventional Resources Technology Conference}, \bibinfo{organization}{URTEC}. p. \bibinfo{pages}{D021S022R002}.
\newblock \DOIprefix\doi{10.15530/urtec-2022-3722510}.
\bibitem[{Jahani et~al.(2023)Jahani, Alyaev, Amb{\'\i}a, Fossum, Suter and Torres-Verd{\'\i}n}]{jahani2023enhancing}
\bibinfo{author}{Jahani, N.}, \bibinfo{author}{Alyaev, S.}, \bibinfo{author}{Amb{\'\i}a, J.}, \bibinfo{author}{Fossum, K.}, \bibinfo{author}{Suter, E.}, \bibinfo{author}{Torres-Verd{\'\i}n, C.}, \bibinfo{year}{2023}.
\newblock \bibinfo{title}{Enhancing the detectability of deep-sensing borehole electromagnetic instruments by joint inversion of multiple logs within a probabilistic geosteering workflow}.
\newblock \bibinfo{journal}{Petrophysics} \bibinfo{volume}{64}, \bibinfo{pages}{80--91}.
\bibitem[{Jahani et~al.(2022)Jahani, Garrido, Alyaev, Fossum, Suter and Torres-Verdín}]{jahani2022}
\bibinfo{author}{Jahani, N.}, \bibinfo{author}{Garrido, J.A.}, \bibinfo{author}{Alyaev, S.}, \bibinfo{author}{Fossum, K.}, \bibinfo{author}{Suter, E.}, \bibinfo{author}{Torres-Verdín, C.}, \bibinfo{year}{2022}.
\newblock \bibinfo{title}{Ensemble-based well-log interpretation and uncertainty quantification for well geosteering}.
\newblock \bibinfo{journal}{GEOPHYSICS} \bibinfo{volume}{87}, \bibinfo{pages}{IM57--IM66}.
\newblock \DOIprefix\doi{10.1190/geo2021-0151.1}.
\bibitem[{Jin et~al.(2020)Jin, Shen, Wu, Chen and Huang}]{jin2020}
\bibinfo{author}{Jin, Y.}, \bibinfo{author}{Shen, Q.}, \bibinfo{author}{Wu, X.}, \bibinfo{author}{Chen, J.}, \bibinfo{author}{Huang, Y.}, \bibinfo{year}{2020}.
\newblock \bibinfo{title}{{A Physics-Driven Deep-Learning Network for Solving Nonlinear Inverse Problems}}.
\newblock \bibinfo{journal}{Petrophysics - The SPWLA Journal of Formation Evaluation and Reservoir Description} \bibinfo{volume}{61}, \bibinfo{pages}{86--98}.
\newblock \DOIprefix\doi{10.30632/PJV61N1-2020a3}.
\bibitem[{Kullawan et~al.(2016a)Kullawan, Bratvold and Bickel}]{Kullawan2014-2}
\bibinfo{author}{Kullawan, K.}, \bibinfo{author}{Bratvold, R.}, \bibinfo{author}{Bickel, J.}, \bibinfo{year}{2016}a.
\newblock \bibinfo{title}{Value creation with multi-criteria decision making in geosteering operations}.
\newblock \bibinfo{journal}{SPE Hydrocarbon Economics and Evaluation Symposium} \DOIprefix\doi{10.2118/169849-MS}.
\bibitem[{Kullawan et~al.(2018)Kullawan, Bratvold and Bickel}]{KULLAWAN201890}
\bibinfo{author}{Kullawan, K.}, \bibinfo{author}{Bratvold, R.}, \bibinfo{author}{Bickel, J.}, \bibinfo{year}{2018}.
\newblock \bibinfo{title}{Sequential geosteering decisions for optimization of real-time well placement}.
\newblock \bibinfo{journal}{Journal of Petroleum Science and Engineering} \bibinfo{volume}{165}, \bibinfo{pages}{90--104}.
\newblock \DOIprefix\doi{10.1016/j.petrol.2018.01.068}.
\bibitem[{Kullawan et~al.(2014)Kullawan, Bratvold and Bickel}]{Kullawan2014}
\bibinfo{author}{Kullawan, K.}, \bibinfo{author}{Bratvold, R.}, \bibinfo{author}{Bickel, J.E.}, \bibinfo{year}{2014}.
\newblock \bibinfo{title}{{A Decision Analytic Approach to Geosteering Operations}}.
\newblock \bibinfo{journal}{SPE Drilling \& Completion} \bibinfo{volume}{29}, \bibinfo{pages}{36--46}.
\newblock \DOIprefix\doi{10.2118/167433-PA}.
\bibitem[{Kullawan et~al.(2016b)Kullawan, Bratvold and Nieto}]{Kullawan2016}
\bibinfo{author}{Kullawan, K.}, \bibinfo{author}{Bratvold, R.B.}, \bibinfo{author}{Nieto, C.M.}, \bibinfo{year}{2016}b.
\newblock \bibinfo{title}{{Decision-Oriented Geosteering and the Value of Look-Ahead Information: A Case-Based Study}}.
\newblock \bibinfo{journal}{SPE Journal} \bibinfo{volume}{22}, \bibinfo{pages}{767--782}.
\newblock \DOIprefix\doi{10.2118/184392-PA}.
\bibitem[{Lu et~al.(2019)Lu, Shen, Chen, Wu and Fu}]{LU2019}
\bibinfo{author}{Lu, H.}, \bibinfo{author}{Shen, Q.}, \bibinfo{author}{Chen, J.}, \bibinfo{author}{Wu, X.}, \bibinfo{author}{Fu, X.}, \bibinfo{year}{2019}.
\newblock \bibinfo{title}{Parallel multiple-chain dram mcmc for large-scale geosteering inversion and uncertainty quantification}.
\newblock \bibinfo{journal}{Journal of Petroleum Science and Engineering} \bibinfo{volume}{174}, \bibinfo{pages}{189--200}.
\newblock \DOIprefix\doi{https://doi.org/10.1016/j.petrol.2018.11.011}.
\bibitem[{Miner et~al.(2021)Miner, Kuvaev and Alyaev}]{log}
\bibinfo{author}{Miner, D.}, \bibinfo{author}{Kuvaev, I.}, \bibinfo{author}{Alyaev, S.}, \bibinfo{year}{2021}.
\newblock \bibinfo{title}{{The typelog from the Geosteering World Cup 2020 semi-finals}}.
\newblock \DOIprefix\doi{10.18710/20VIVT}.
\bibitem[{Mnih et~al.(2015)Mnih, Kavukcuoglu, Silver, Rusu, Veness, Bellemare, Graves, Riedmiller, Fidjeland, Ostrovski, Petersen, Beattie, Sadik, Antonoglou, King, Kumaran, Wierstra, Legg and Hassabis}]{Mnih2015-mf}
\bibinfo{author}{Mnih, V.}, \bibinfo{author}{Kavukcuoglu, K.}, \bibinfo{author}{Silver, D.}, \bibinfo{author}{Rusu, A.A.}, \bibinfo{author}{Veness, J.}, \bibinfo{author}{Bellemare, M.G.}, \bibinfo{author}{Graves, A.}, \bibinfo{author}{Riedmiller, M.}, \bibinfo{author}{Fidjeland, A.K.}, \bibinfo{author}{Ostrovski, G.}, \bibinfo{author}{Petersen, S.}, \bibinfo{author}{Beattie, C.}, \bibinfo{author}{Sadik, A.}, \bibinfo{author}{Antonoglou, I.}, \bibinfo{author}{King, H.}, \bibinfo{author}{Kumaran, D.}, \bibinfo{author}{Wierstra, D.}, \bibinfo{author}{Legg, S.}, \bibinfo{author}{Hassabis, D.}, \bibinfo{year}{2015}.
\newblock \bibinfo{title}{Human-level control through deep reinforcement learning}.
\newblock \bibinfo{journal}{Nature} \bibinfo{volume}{518}, \bibinfo{pages}{529--533}.
\bibitem[{Muhammad et~al.(2023)Muhammad, Alyaev and Bratvold}]{muhammad2023optimal}
\bibinfo{author}{Muhammad, R.B.}, \bibinfo{author}{Alyaev, S.}, \bibinfo{author}{Bratvold, R.B.}, \bibinfo{year}{2023}.
\newblock \bibinfo{title}{Optimal sequential decision-making in geosteering: A reinforcement learning approach}.
\newblock \href{http://arxiv.org/abs/2310.04772}{{\tt arXiv:2310.04772}}.
\bibitem[{Paszke et~al.(2019)Paszke, Gross, Massa, Lerer, Bradbury, Chanan, Killeen, Lin, Gimelshein, Antiga, Desmaison, Kopf, Yang, DeVito, Raison, Tejani, Chilamkurthy, Steiner, Fang, Bai and Chintala}]{pytorch}
\bibinfo{author}{Paszke, A.}, \bibinfo{author}{Gross, S.}, \bibinfo{author}{Massa, F.}, \bibinfo{author}{Lerer, A.}, \bibinfo{author}{Bradbury, J.}, \bibinfo{author}{Chanan, G.}, \bibinfo{author}{Killeen, T.}, \bibinfo{author}{Lin, Z.}, \bibinfo{author}{Gimelshein, N.}, \bibinfo{author}{Antiga, L.}, \bibinfo{author}{Desmaison, A.}, \bibinfo{author}{Kopf, A.}, \bibinfo{author}{Yang, E.}, \bibinfo{author}{DeVito, Z.}, \bibinfo{author}{Raison, M.}, \bibinfo{author}{Tejani, A.}, \bibinfo{author}{Chilamkurthy, S.}, \bibinfo{author}{Steiner, B.}, \bibinfo{author}{Fang, L.}, \bibinfo{author}{Bai, J.}, \bibinfo{author}{Chintala, S.}, \bibinfo{year}{2019}.
\newblock \bibinfo{title}{Pytorch: An imperative style, high-performance deep learning library}, in: \bibinfo{booktitle}{Advances in Neural Information Processing Systems 32}. \bibinfo{publisher}{Curran Associates, Inc.}, pp. \bibinfo{pages}{8024--8035}.
\bibitem[{Puterman(1990)}]{PUTERMAN1990331}
\bibinfo{author}{Puterman, M.L.}, \bibinfo{year}{1990}.
\newblock \bibinfo{title}{Chapter 8 markov decision processes}, in: \bibinfo{booktitle}{Stochastic Models}. \bibinfo{publisher}{Elsevier}. volume~\bibinfo{volume}{2} of \textit{\bibinfo{series}{Handbooks in Operations Research and Management Science}}, pp. \bibinfo{pages}{331--434}.
\newblock \DOIprefix\doi{https://doi.org/10.1016/S0927-0507(05)80172-0}.
\bibitem[{Puzyrev and Swidinsky(2021)}]{puzyrev2021inversion}
\bibinfo{author}{Puzyrev, V.}, \bibinfo{author}{Swidinsky, A.}, \bibinfo{year}{2021}.
\newblock \bibinfo{title}{Inversion of 1d frequency-and time-domain electromagnetic data with convolutional neural networks}.
\newblock \bibinfo{journal}{Computers \& geosciences} \bibinfo{volume}{149}, \bibinfo{pages}{104681}.
\bibitem[{Rammay et~al.(2022)Rammay, Alyaev and Elsheikh}]{rammay2022probabilistic}
\bibinfo{author}{Rammay, M.H.}, \bibinfo{author}{Alyaev, S.}, \bibinfo{author}{Elsheikh, A.H.}, \bibinfo{year}{2022}.
\newblock \bibinfo{title}{Probabilistic model-error assessment of deep learning proxies: an application to real-time inversion of borehole electromagnetic measurements}.
\newblock \bibinfo{journal}{Geophysical Journal International} \bibinfo{volume}{230}, \bibinfo{pages}{1800--1817}.
\bibitem[{Ristic et~al.(2004)Ristic, Arulampalam and Gordon}]{PFristic}
\bibinfo{author}{Ristic, B.}, \bibinfo{author}{Arulampalam, S.}, \bibinfo{author}{Gordon, N.}, \bibinfo{year}{2004}.
\newblock \bibinfo{title}{Beyond the Kalman Filter: Particle Filters for Tracking Applications}.
\newblock Artech House radar library, \bibinfo{publisher}{Artech House}.
\newblock \URLprefix \url{https://books.google.com/books?id=cjFDngEACAAJ}.
\bibitem[{Shahriari et~al.(2020)Shahriari, Pardo, Pic{\'o}n, Galdran, Del~Ser and Torres-Verd{\'\i}n}]{shahriari2020deep}
\bibinfo{author}{Shahriari, M.}, \bibinfo{author}{Pardo, D.}, \bibinfo{author}{Pic{\'o}n, A.}, \bibinfo{author}{Galdran, A.}, \bibinfo{author}{Del~Ser, J.}, \bibinfo{author}{Torres-Verd{\'\i}n, C.}, \bibinfo{year}{2020}.
\newblock \bibinfo{title}{A deep learning approach to the inversion of borehole resistivity measurements}.
\newblock \bibinfo{journal}{Computational Geosciences} \bibinfo{volume}{24}, \bibinfo{pages}{971--994}.
\bibitem[{Shahriari et~al.(2021)Shahriari, Pardo, Rivera, Torres-Verd{\'\i}n, Picon, Del~Ser, Ossandon and Calo}]{shahriari2021error}
\bibinfo{author}{Shahriari, M.}, \bibinfo{author}{Pardo, D.}, \bibinfo{author}{Rivera, J.A.}, \bibinfo{author}{Torres-Verd{\'\i}n, C.}, \bibinfo{author}{Picon, A.}, \bibinfo{author}{Del~Ser, J.}, \bibinfo{author}{Ossandon, S.}, \bibinfo{author}{Calo, V.M.}, \bibinfo{year}{2021}.
\newblock \bibinfo{title}{Error control and loss functions for the deep learning inversion of borehole resistivity measurements}.
\newblock \bibinfo{journal}{International Journal for Numerical Methods in Engineering} \bibinfo{volume}{122}, \bibinfo{pages}{1629--1657}.
\bibitem[{Shen et~al.(2018)Shen, Wu, Chen, Han and Huang}]{SHEN2018}
\bibinfo{author}{Shen, Q.}, \bibinfo{author}{Wu, X.}, \bibinfo{author}{Chen, J.}, \bibinfo{author}{Han, Z.}, \bibinfo{author}{Huang, Y.}, \bibinfo{year}{2018}.
\newblock \bibinfo{title}{Solving geosteering inverse problems by stochastic hybrid monte carlo method}.
\newblock \bibinfo{journal}{Journal of Petroleum Science and Engineering} \bibinfo{volume}{161}, \bibinfo{pages}{9--16}.
\newblock \DOIprefix\doi{10.1016/j.petrol.2017.11.031}.
\bibitem[{Srivastava et~al.(2023)Srivastava, Kang and Tartakovsky}]{SRIVASTAVA2023112499}
\bibinfo{author}{Srivastava, A.}, \bibinfo{author}{Kang, W.}, \bibinfo{author}{Tartakovsky, D.M.}, \bibinfo{year}{2023}.
\newblock \bibinfo{title}{Feature-informed data assimilation}.
\newblock \bibinfo{journal}{Journal of Computational Physics} \bibinfo{volume}{494}, \bibinfo{pages}{112499}.
\newblock \URLprefix \url{https://www.sciencedirect.com/science/article/pii/S0021999123005946}, \DOIprefix\doi{https://doi.org/10.1016/j.jcp.2023.112499}.
\bibitem[{Sutton and Barto(2018)}]{sutton2018}
\bibinfo{author}{Sutton, R.S.}, \bibinfo{author}{Barto, A.G.}, \bibinfo{year}{2018}.
\newblock \bibinfo{title}{Reinforcement Learning: An Introduction}.
\newblock \bibinfo{publisher}{A Bradford Book}, \bibinfo{address}{Cambridge, MA, USA}.
\bibitem[{Sviridov et~al.(2014)Sviridov, Mosin, Antonov, Nikitenko, Martakov and Rabinovich}]{sviridov2014new}
\bibinfo{author}{Sviridov, M.}, \bibinfo{author}{Mosin, A.}, \bibinfo{author}{Antonov, Y.}, \bibinfo{author}{Nikitenko, M.}, \bibinfo{author}{Martakov, S.}, \bibinfo{author}{Rabinovich, M.}, \bibinfo{year}{2014}.
\newblock \bibinfo{title}{New software for processing of lwd extradeep resistivity and azimuthal resistivity data}.
\newblock \bibinfo{journal}{SPE Reservoir Evaluation \& Engineering} \bibinfo{volume}{17}, \bibinfo{pages}{109--127}.
\bibitem[{Tadjer et~al.(2021)Tadjer, Alyaev, Miner, Kuvaev and Bratvold}]{amine2021}
\bibinfo{author}{Tadjer, A.}, \bibinfo{author}{Alyaev, S.}, \bibinfo{author}{Miner, D.}, \bibinfo{author}{Kuvaev, I.}, \bibinfo{author}{Bratvold, R.B.}, \bibinfo{year}{2021}.
\newblock \bibinfo{title}{{Unlocking the Human Factor: Geosteering Decision Making as a Component of Drilling Operational Efficacy}} \bibinfo{volume}{Day 2 Tue, July 27, 2021}, \bibinfo{pages}{D021S028R001}.
\newblock \DOIprefix\doi{10.15530/urtec-2021-5385}.
\bibitem[{Thiel et~al.(2018)Thiel, Bower and Omeragic}]{thiel20182d}
\bibinfo{author}{Thiel, M.}, \bibinfo{author}{Bower, M.}, \bibinfo{author}{Omeragic, D.}, \bibinfo{year}{2018}.
\newblock \bibinfo{title}{2d reservoir imaging using deep directional resistivity measurements}.
\newblock \bibinfo{journal}{Petrophysics} \bibinfo{volume}{59}, \bibinfo{pages}{218--233}.
\bibitem[{Tsitsiklis and Van~Roy(1997)}]{TDdivergence}
\bibinfo{author}{Tsitsiklis, J.}, \bibinfo{author}{Van~Roy, B.}, \bibinfo{year}{1997}.
\newblock \bibinfo{title}{An analysis of temporal-difference learning with function approximation}.
\newblock \bibinfo{journal}{IEEE Transactions on Automatic Control} \bibinfo{volume}{42}, \bibinfo{pages}{674--690}.
\newblock \DOIprefix\doi{10.1109/9.580874}.
\bibitem[{Veettil and Clark(2020)}]{Veettil20}
\bibinfo{author}{Veettil, D.R.A.}, \bibinfo{author}{Clark, K.}, \bibinfo{year}{2020}.
\newblock \bibinfo{title}{{Bayesian Geosteering Using Sequential Monte Carlo Methods}}.
\newblock \bibinfo{journal}{Petrophysics - The SPWLA Journal of Formation Evaluation and Reservoir Description} \bibinfo{volume}{61}, \bibinfo{pages}{99--111}.
\newblock \DOIprefix\doi{10.30632/PJV61N1-2020a4}.
\bibitem[{Vetra-Carvalho et~al.(2018)Vetra-Carvalho, van Leeuwen, Nerger, Barth, Altaf, Brasseur, Kirchgessner and Beckers}]{BayesianDAReview}
\bibinfo{author}{Vetra-Carvalho, S.}, \bibinfo{author}{van Leeuwen, P.J.}, \bibinfo{author}{Nerger, L.}, \bibinfo{author}{Barth, A.}, \bibinfo{author}{Altaf, M.U.}, \bibinfo{author}{Brasseur, P.}, \bibinfo{author}{Kirchgessner, P.}, \bibinfo{author}{Beckers, J.M.}, \bibinfo{year}{2018}.
\newblock \bibinfo{title}{State-of-the-art stochastic data assimilation methods for high-dimensional non-gaussian problems}.
\newblock \bibinfo{journal}{Tellus A: Dynamic Meteorology and Oceanography} \bibinfo{volume}{70}, \bibinfo{pages}{1--43}.
\newblock \DOIprefix\doi{10.1080/16000870.2018.1445364}.
\bibitem[{Wu et~al.(2018)Wu, Golla, Parker, Clegg and Monteilhet}]{wu2018new}
\bibinfo{author}{Wu, H.H.}, \bibinfo{author}{Golla, C.}, \bibinfo{author}{Parker, T.}, \bibinfo{author}{Clegg, N.}, \bibinfo{author}{Monteilhet, L.}, \bibinfo{year}{2018}.
\newblock \bibinfo{title}{A new ultra-deep azimuthal electromagnetic lwd sensor for reservoir insight}, in: \bibinfo{booktitle}{SPWLA Annual Logging Symposium}, \bibinfo{organization}{SPWLA}. p. \bibinfo{pages}{D043S007R004}.

\end{thebibliography}





\end{document}